\documentclass{article} 
\usepackage{iclr2027_conference,times}

\usepackage{amsmath,amsfonts,bm}

\def\eqref#1{equation~\ref{#1}}

\def\1{\bm{1}}

\def\vs{{\bm{s}}}

\def\vu{{\bm{u}}}

\def\mA{{\bm{A}}}

\def\mG{{\bm{G}}}

\def\mL{{\bm{L}}}
\def\mM{{\bm{M}}}

\def\mU{{\bm{U}}}

\def\mW{{\bm{W}}}
\def\mX{{\bm{X}}}
\def\mY{{\bm{Y}}}

\DeclareMathAlphabet{\mathsfit}{\encodingdefault}{\sfdefault}{m}{sl}
\SetMathAlphabet{\mathsfit}{bold}{\encodingdefault}{\sfdefault}{bx}{n}

\usepackage{hyperref}
\usepackage{url}
\usepackage{wrapfig}
\usepackage{tabularx}
\usepackage{booktabs}
\usepackage{graphicx}
\usepackage{multirow}
\usepackage[table]{xcolor}
\usepackage{array}
\usepackage{caption}

\usepackage{listings}
\usepackage{xcolor}

\usepackage{xurl}
\usepackage{hyperref}

\title{SpecMRL: Redesign Matryoshka Representation Learning within Parametric Spectrum}

\author{
  Zhongxin Chen$^{1}$\thanks{These authors contributed equally. Zhongxin Chen was responsible for the design, implementation, optimization, and evaluation of \name. Songhao Wu proposed the initial idea of \name\ and led this project.}, \; 
  Songhao Wu$^{1}$\footnotemark[1], \; 
  Heng Cui$^{3}$, \;
  Cong Li$^{3}$, \; 
  Xiao Zhou$^{1}$\thanks{Corresponding authors.}, \; 
  Rui Yan$^{2}$\footnotemark[2] \\
  $^{1}$ Gaoling School of Artificial Intelligence, Renmin University of China \\
  $^{2}$ Lenovo Group \quad $^{3}$ School of Computer Science, Wuhan University   \\ 
  \small \texttt{\{chenzhongxin, songhaowu, xiaozhou\}@ruc.edu.cn} \; \texttt{rui.yan@whu.edu.cn}
}

\newcommand{\name}{SpecMRL}
\definecolor{gridblue}{RGB}{45, 110, 220}
\newcommand{\mrlweight}[1]{%
    \colorbox{blue!10}{\strut $(#1)$}%
}
\newcommand{\padzero}{\textcolor{white}{0}}
\newcommand{\hflink}[1]{%
    \href{#1}{%
        \raisebox{-0.15em}{\includegraphics[height=1.15em]{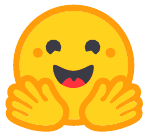}}%
    }%
}

\iclrfinalcopy 
\begin{document}

\maketitle
\lhead{Preprint.}

\begin{abstract}
Matryoshka Representation Learning (MRL) is designed to support flexible representation sizes by jointly optimizing nested prefixes.
While effective, this design lacks an explicit constraint to ensure that the leading portion provides a suitable compact encoding.
To address this, we propose \name, which reorganizes the feature space to enable more effective prefix readout.
Specifically, we extend MRL with a parameter-derived spectral basis organized with feature importance.
\name\ otherwise follows standard MRL, with prefix readout performed in the learned spectral space.
Experiments on MTEB show that \name\ outperforms MRL at all evaluated reduced dimensions, with larger gains under more aggressive compression.
These results suggest that the learned spectral basis better prioritizes informative directions, enabling more effective representation compression.
Notably, \name\ extends the performance frontier of MRL, surpassing full-dimensional MRL with only half the embedding dimensions.\footnote{Our code is available at \url{https://github.com/CentreChen/spec-mrl}.}
\end{abstract}

\section{Introduction}
Representation learning has been a well-established research topic in the ML community~\citep{simclr,simcse} and underpins a wide range of downstream applications~\citep{sbert,dpr}.
The growing applications call for more flexible representation schemes that can adapt to varying latency and storage constraints across use cases.
Matryoshka Representation Learning (MRL, \citealp{mrl}) was introduced to address this need, offering an approach that can be easily adopted in existing representation learning frameworks. 

The central idea of MRL is to extend full-dimensional representation learning to a set of nested prefixes.
In practice, each nested readout is constructed from the leading portion of the full representation.
This design enables representations at multiple output dimensions to be learned within a single training run.
While simple and effective, it does not explicitly ensure that earlier dimensions are better suited for nested readout, motivating a more principled design for MRL.

\begin{figure}[t]
    \centering
    \includegraphics[width=\linewidth]{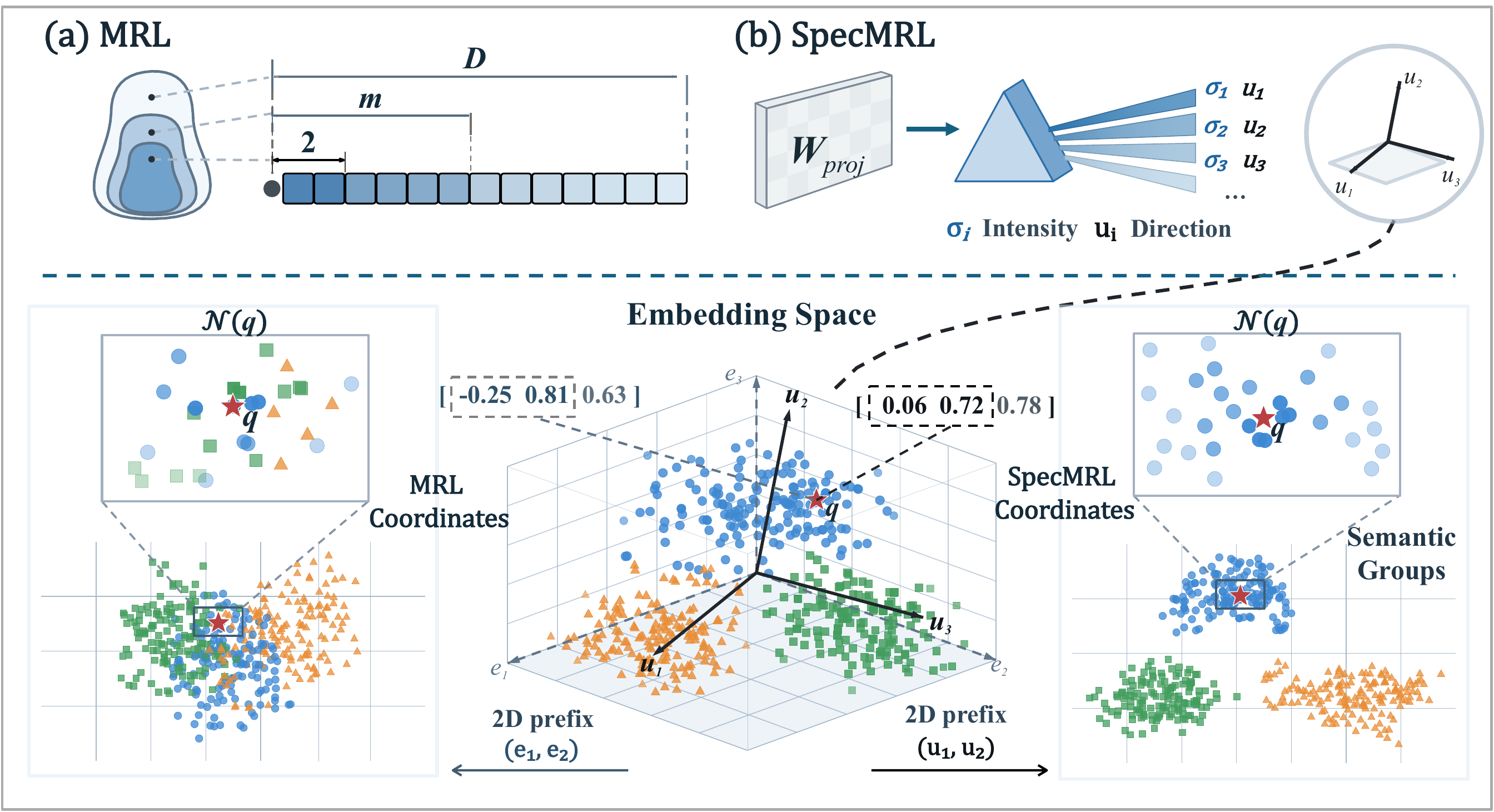}
    \caption{
    \textbf{Schematic comparison of standard MRL and \name.}
    Text embeddings are shown as points in the middle, with colors and marker shapes denoting distinct semantic groups.
    MRL retains coordinate prefixes in the original basis $\{e_i\}$ as compact representations, whereas \name\ performs the readout in an ordered spectral basis $\{u_i\}$ derived from the parameter structure of $\mW_{\mathrm{Proj}}$.
    The 2D views and enlarged neighborhoods of $q$ (red star) illustrate how this change better organizes the semantic space and supports more effective low-dimensional representations.
    }
    \vspace{-12pt}
    \label{fig:teaser}
\end{figure}

For such prefix readout to be effective, the leading dimensions should ideally concentrate the information most useful for compact representations.
However, standard MRL provides no explicit constraint for establishing this ordering.
Earlier studies~\citep{ese,mipic} have sought to improve low-dimensional readouts through cross-dimensional supervision.
Yet, these approaches keep the original coordinate system unchanged, which may itself be ill-suited to effective prefix truncation.
Figure~\ref{fig:teaser} depicts how a more suitable basis can better organize semantic information and improve the quality of low-dimensional representations.
This motivates us to rethink how the representation space itself should be organized to better support nested prefix readout in MRL.

To this end, we propose \name, a parameter-derived spectral redesign of MRL that organizes the representation space by feature importance.
Such an organization better aligns with the prefix-based design of MRL and provides a natural foundation for dimension reduction.
Starting from a backbone representation, \name\ differs from standard MRL only in its choice of basis.
Figure~\ref{fig:teaser} presents a comparative overview of MRL and \name, highlighting different organizations of representation space for nested readout.
Specifically, \name\ changes only the representation basis, with standard MRL subsequently applied in the resulting space.


Concretely, we introduce a matrix $\mW_{\mathrm{Proj}}$ and derive from its Gram matrix an ordered spectral basis that serves as the desired coordinate system.
Although this is simple in design, a direct implementation poses two challenges: (1) the computational overhead of standard eigen solvers and (2) the numerical instability of eigen vector gradients.
To address these challenges, we introduce several technical refinements and develop a power-iteration-based alternative to approximate the eigen decomposition of the Gram matrix.
Our empirical observations further confirm that this design avoids training collapse due to numerical instability while achieving throughput comparable to vanilla MRL.

We conduct experiments to validate the effectiveness of \name\ and provide further insights into its design.
Specifically, we evaluate \name\ on widely used MTEB datasets~\citep{mteb} with backbones from multiple model families.
The empirical results show that \name\ substantially improves performance at reduced dimensions, outperforming MRL by up to 3.80 points under $16\times$ compression.
Beyond these gains, \name\ further extends the performance frontier of MRL, surpassing its full-capacity performance with only half the dimensions.
Further validation shows that these gains persist with larger backbones and longer training, confirming the potential of its further scaling.
Our analyses also provide initial insights into why \name\ is effective.
Taken together, these results motivate a rethink of the design principles of MRL and we hope it can inspire improvements for future representation learning.

\section{Background: Matryoshka Representation Learning (MRL)}
We briefly review how sentence embeddings are obtained and trained with LLMs before the introduction of MRL~\citep{mrl}.
Given an input sequence $\vs$, the LLM encodes it into contextualized hidden states and aggregates them into a dense representation $\mX$:
\[
    \mX = \mathcal{\mathbf{P}}\left(\operatorname{\mathrm{LLM}}\left([\vs_1, \vs_2, \dots, \vs_n]\right)\right).
\]
Given a training batch $\mathcal{B}$, $\mX$ is typically optimized via the following contrastive loss:
\begin{equation}
    \mathcal{L}\left(\mX\right) = -\frac{1}{|\mathcal{B}|}\sum_{\mX \in \mathcal{B}}
    \log\frac{\exp\big(\operatorname{sim}(\mX, \mX^+; T)\big)}{\exp\big(\operatorname{sim}(\mX, \mX^+;T)\big) + \sum_{\mX^- \in \mathcal{N}}\exp\big(\operatorname{sim}(\mX, \mX^-; T)\big)
    },
\end{equation}
where $\mX^+$ and $\mX^-$ denote the positive and negative samples of $\mX$, and $T$ controls the temperature of the similarity function $\operatorname{sim}(\cdot,\cdot\,;\,T)$.
MRL extends this contrastive learning over a set of nested dimensions $\mathcal{M}$, with loss applied at each dimension.
Specifically, for any nested dimension $m \in \mathcal{M}$, MRL retains the first $m$ entries of $\mX$ as the reduced representation:
\begin{equation}
    \mX_{\textrm{MRL}}^{(m)} \, = \, \mX[:\,, \,:m],
\end{equation}
and aggregates the losses across nested dimensions into a weighted objective:
\begin{equation}
    \mathcal{L}_{\mathrm{MRL}} \;=\; \sum\nolimits_{m \in \mathcal{M}} w_m \, \mathcal{L}\left(\mX^{(m)}_{\textrm{MRL}}\right) \;,
\end{equation}
where $w_m$ denotes the weight assigned to each dimension $m$.
This nested structure motivates the name, \underline{M}atryoshka \underline{R}epresentation \underline{L}earning~(\textbf{MRL}).
Here, the “Matryoshka” metaphor is motivated by Russian nesting dolls, reflecting the intuition that more informative features should occupy earlier dimensions.
Despite this motivation, MRL does not explicitly ensure such an ordering, leaving the distribution of information across dimensions unspecified.
Prior work~\citep{ese,adawidth,smec} has also considered this issue, but mainly introduces additional constraints across nested representations instead of directly ordering the feature dimensions.
In this paper, we seek to establish refined design principles toward better dimension reduction.
Below, we present \name\ as a refined perspective on how the core motivation behind MRL can be realized.

\section{Methodology}
\label{sec:method}
\subsection{\name: Parametric Spectral Design for Improved MRL}
In this section, we first present the implementation of \name\ and defer its design principles to later sections.
By design, MRL relies on a heuristic prior that favors earlier dimensions to construct nested representations.
\name\ breaks this paradigm and recasts MRL's predefined dimension ordering as a learnable spectral organization.
Specifically, it encodes this latent structure in $\mW_{\mathrm{Proj}} \in \mathbb{R}^{D \times R}$ and recovers an ordered basis from the spectrum of the corresponding Gram matrix:
\begin{equation}
    \mU \, = \, \operatorname{\mathbf{Eigen}}\!\left(\,\mW_{\mathrm{Proj}} \, \mW_{\mathrm{Proj}}^\top\,\right),
    \label{eq:eigh}
\end{equation}
where $\mathbf{Eigen}(\cdot)$ returns the basis $\mU$, with eigen values ordered as $\lambda_1 \geq \lambda_2 \geq \cdots \geq \lambda_D \geq 0$.
Keeping the leading $m$ columns
$\mU_{:,1:m} \in \mathbb{R}^{D\times m}$ as the nested
readout basis, we define
\begin{equation}
    \mX_{\mathrm{Spec}}^{(m)}
    =
    \mX \mU_{:,1:m}
    \in \mathbb{R}^{m}
\end{equation}
where \(\mX^{(m)}_{\textrm{Spec}}\) denotes the projection of $\mX \in \mathbb{R}^D$ onto the leading $m$ basis vectors.
With this definition, \name\ is a drop-in modification to MRL, replacing the nested representation $\mX^{(m)}$ with the spectral prefix $\mX^{(m)}_{\textrm{Spec}}$ in each loss term while leaving the rest of the pipeline unchanged.

\begin{figure*}[t]
    \centering

    \begin{minipage}[b]{0.47\textwidth}
        \centering
        \includegraphics[width=\linewidth]{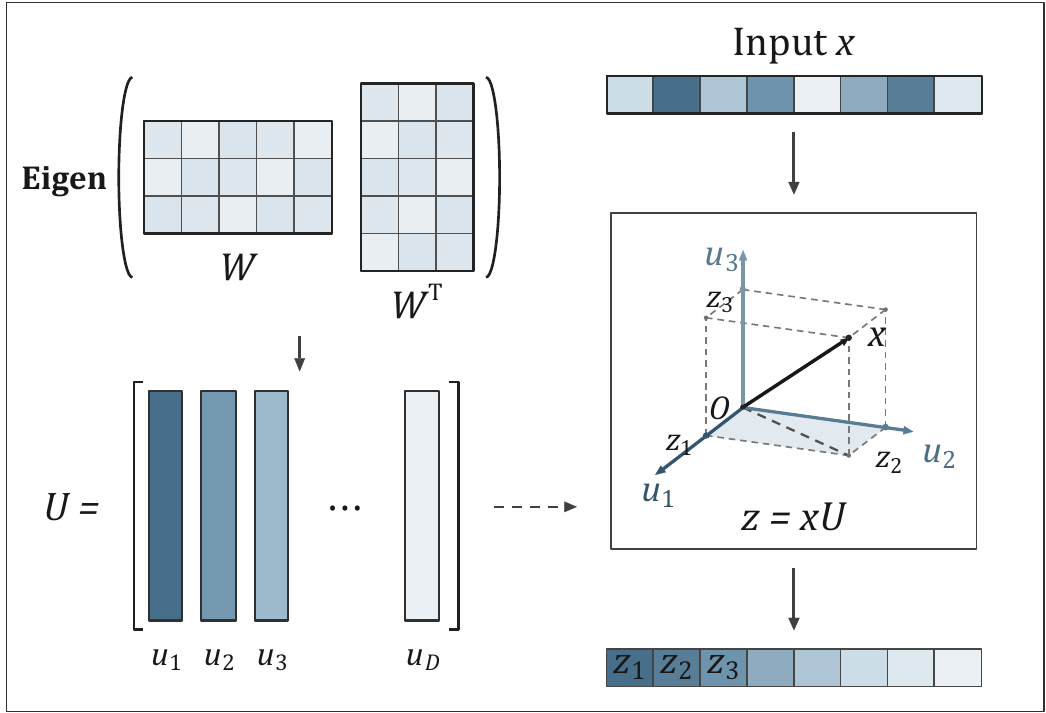}
        \par\vspace{3pt}
        {\small (a) Spectral readout.\par}
    \end{minipage}
    \hfill
    \begin{minipage}[b]{0.48\textwidth}
        \definecolor{codegreen}{rgb}{0.0,0.4,0.0}
\begin{lstlisting}[
    language=Python,
    basicstyle=\linespread{1}\fontfamily{qcr}
               \fontsize{7.8}{10}\selectfont\color{black},
    keywordstyle=\color{blue!75!black}\bfseries,
    commentstyle=\color{codegreen},
    stringstyle=\color{red},
    frame=tb,
    rulecolor=\color{black},
    framerule=0.4pt,
    framesep=4pt,
    aboveskip=0pt,
    belowskip=0pt,
    numbers=none,
    tabsize=4,
    keepspaces=true,
    breaklines=false,
    showstringspaces=false,
    columns=flexible,
    basewidth={0.5em,0.42em}
]
# W: learnable factor [D, R]
# U: cached spectral basis [D, D]
def specmrl(q, p, W, U, y):
    # Warm-started spectral update
    G = W @ W.T
    U, _ = torch.linalg.qr(G @ U.detach())
    # Spectral readout
    q, p = q @ U, p @ U
    
    # Original Matryoshka loss
    loss = mrl_loss(q, p, y)
    return loss, U.detach()
\end{lstlisting}
        \par\vspace{5pt}
        {\centering\small (b) PyTorch-style pseudocode.\par}
    \end{minipage}

    \caption{
        Illustration and pseudocode of \name.
        (a) Embeddings are rotated into a learned spectral
        basis before prefix readouts.
        (b) The basis is updated by one warm-started
        orthogonal iteration per training step and cached
        for the next step.
        \texttt{mrl\_loss} denotes the original Matryoshka objective.
    }
    \vspace{-12pt}
    \label{fig:specmrl}
\end{figure*}

Given the constructed spectrum, the training objective of \name\ can be formulated as:
\begin{equation}
    \mathcal{L}_{\mathrm{Spec}} \;=\; \sum\nolimits_{m \in \mathcal{M}} w_m \, \mathcal{L}\left(\mX^{(m)}_{\textrm{Spec}}\right) \;.
    \label{eq:mrl}
\end{equation}
This substitution changes only the construction of nested representations while preserving the vanilla MRL framework, and requires no additional loss terms or modifications to the training pipeline.
Figure~\ref{fig:specmrl} provides an overview of the key stages in the \name\ pipeline.

We present a parameter-space view to help interpret our \name\ design.
Conceptually, the learned parameters of $\mW_{\mathrm{Proj}}$ can be viewed as an encoding of the embedding-space geometry.
From this perspective, \name\ may organize feature channels in a more principled manner than vanilla MRL, offering one possible explanation for the observed performance gains.
We provide additional empirical observations in Appendix~\ref{app:parameter-geometry} to further support this interpretation.

\subsection{\name\ with Efficient and Scalable Spectral Solver}
While \name\ is simple in design, its naive implementation can involve several subtle technical challenges that are easy to overlook.
A straightforward approach is to directly decompose the Gram matrix using off-the-shelf numerical tools, such as the built-in eigen solvers in PyTorch~\citep{pytorch}, to obtain the spectral basis.
However, this is (1) computationally expensive and (2) sensitive to ill-conditioned eigenvalue distribution, making it unsuitable in practice.\footnote{For readers unfamiliar with this issue, we provide a detailed discussion in Section~\ref{sec:efficiency}.}

To address these challenges, we approximate the ordered basis in Eq.~\ref{eq:eigh} with a streaming power-iteration design.
Specifically, we first initialize the basis with the ordered eigen vectors of the initial Gram matrix.
Starting from this initialization, each subsequent step updates the previous basis through a single matrix multiplication followed by QR orthogonalization~\citep{qr}:
\begin{equation}
    \mU^{(t)}
    =
    \operatorname{QR}\!\left(
        \mW_{\mathrm{Proj}}^{(t)}
        \mW_{\mathrm{Proj}}^{(t)\top}
        \mU^{(t-1)}
    \right),
\end{equation}
where $\operatorname{QR}$ returns the orthogonal factor.
Gradients propagate through the Gram matrix and QR factorization, while basis $\mU^{(t-1)}$ is held fixed.
Similar approaches have been explored in optimizer designs to reduce decomposition overhead~\citep{dion,soap}.
Figure~\ref{fig:specmrl}~(b) provides PyTorch-style pseudocode for this design. 
We further validate the convergence of this solver in Appendix~\ref{app:parameter-geometry} by comparing its solved basis against the exact eigen decomposition solution.

The benefits of our power iteration design are empirically validated in practice.
In particular, our power-iteration design avoids training collapse across all experimental settings with only modest computational overhead.
Once training is complete, the spectral basis is fixed and no longer requires recomputation from $\mW_{\mathrm{Proj}}$. 
As a result, \name\ incurs only the cost of a $D \times D$ linear projection at inference time. 
We provide a detailed analysis of both training and inference costs in Section~\ref{sec:efficiency} to further assess the practical efficiency of \name.

\section{Experiment}

\subsection{Method Analysis: Clarifying Design Principles for \name}
We begin with ablation studies to shed light on the research questions regarding \name, seeking to validate its effectiveness and provide insights into the design principles behind it.

\subsubsection{How does \name\ perform relative to traditional MRL?}
\label{sec:main}
We train \name\ alongside MRL on backbones from various model families and evaluate their performance on MTEB~\citep{mteb}.
Table~\ref{tab:main} summarizes the results, while detailed experimental setups are provided in Appendix~\ref{app:setup}.
\name\ highlights the following advantages:

(1) \name\ outperforms MRL across all reduced dimensions.
The observed improvements are consistent across backbones and compression ratios, while \name\ preserves the nested structure of MRL.
Importantly, the performance gap widens under stronger compression.
The larger gains at lower dimensions suggest that \name\ more effectively prioritizes informative directions for low-dimensional readouts, which may explain its advantage over MRL.

(2) \name\ extends the frontier of MRL, achieving superior performance using a compact representation.
As shown in Table~\ref{tab:main}, \name\ achieves its best performance with only half of the full dimension.
This reduction lowers retrieval costs, with negligible overhead and no compromise in downstream performance.
We further analyze this dimension--performance trade-off, including the slight degradation observed at the full dimension, in Section~\ref{sec:scalabilty}.

We argue that these findings position \name\ as a promising alternative to standard MRL, pushing the performance frontier toward more compact representations.

\begin{table}[t]
    \caption{Performance of \name\ averaged across MTEB tasks, with task-level breakdown provided in {Table~\ref{tab:breakdown}}. $\tau$ controls the reduction ratio, reducing the output dimension to $1/\tau$ of its original size. Colored entries highlight improvements over vanilla MRL, while bold text marks the best result for each backbone. Parenthetical values denote the gains of \name\ over MRL.}
    \label{tab:main}
    \centering
    \small
    \resizebox{0.88\textwidth}{!}{
        \begin{tabular}{c|c|c|c|c|c|c}
            \toprule
            & \multicolumn{2}{c}{\textbf{\textrm{Qwen3-0.6B}}} & \multicolumn{2}{c}{\textbf{\textrm{Qwen2-0.5B}}} & \multicolumn{2}{c}{\textbf{\textrm{Llama3.2-1B}}} \\
            \cmidrule(l{1pt}r{1pt}){1-1} \cmidrule(l{2pt}r{2pt}){2-3} \cmidrule(l{2pt}r{2pt}){4-5} \cmidrule(l{2pt}r{1pt}){6-7}
            $\mathbf{\tau}$ & MRL & \name & MRL & \name & MRL & \name \\
            \midrule
            1 & 60.97 & 60.94~($-$0.03) & 58.96 & 58.74~($-$0.22) & 61.92 & 61.87~($-$0.05) \\
            \cmidrule(l{1pt}r{1pt}){1-1} \cmidrule(l{2pt}r{2pt}){2-3} \cmidrule(l{2pt}r{2pt}){4-5} \cmidrule(l{2pt}r{1pt}){6-7}
            2 & 60.54 & \cellcolor{gray!12}\textbf{61.61}~(+1.07) & 58.67 & \cellcolor{gray!12}\textbf{59.44}~(+0.77) & 61.74 & \cellcolor{gray!12}\textbf{62.42}~(+0.68) \\ [1pt]
            4 & 59.40 & \cellcolor{gray!12}61.28~(+1.88) & 58.32 & \cellcolor{gray!12}59.20~(+0.88) & 61.36 & \cellcolor{gray!12}62.23~(+0.87) \\ [1pt]
            8 & 57.88 & \cellcolor{gray!12}60.46~(+2.58) & 57.26 & \cellcolor{gray!12}58.44~(+1.18) & 60.62 & \cellcolor{gray!12}61.80~(+1.18) \\ [1pt]
            16 & 54.73 & \cellcolor{gray!12}58.53~(+3.80) & 55.23 & \cellcolor{gray!12}56.61~(+1.38) & 59.42 & \cellcolor{gray!12}60.91~(+1.49) \\
            \bottomrule
        \end{tabular}
    }
    \vspace{-8pt}
\end{table}

\subsubsection{What motivates the use of parametric spectrum for \name?}
The spectral basis in \name\ can, in principle, be derived from sources beyond the learned parameter matrix.
Below, we introduce several alternatives to clarify the rationale behind our choice.

\paragraph{Parameter-Derived Spectrum Proves Effective.}
One natural alternative to \name's parameter-derived spectrum is to construct the spectral basis directly from the input features.
Prior work~\citep{ese} has also explored exploiting feature-space structure to guide MRL, making feature-derived basis a plausible choice.
We further discuss the relation to ESE in Section~\ref{sec:ese}.

\begin{table}[b]
    \caption{Comparison of \name\ and FeatMRL on MTEB. Bold values indicate the best performance, while parenthetical values denote performance drops relative to \name.}
    \label{tab:featmrl}
    \centering
    \small
    \resizebox{0.98\textwidth}{!}{
        \begin{tabular}{c|c|c|c|c|c}
            \toprule
            $\mathbf{\tau}$ & \name & FeatMRL $(\beta=0)$ & FeatMRL $(\beta=0.1)$ & FeatMRL $(\beta=0.5)$ & FeatMRL $(\beta=0.9)$ \\
            \midrule
            1  & \textbf{60.94} & 60.72~($-$0.22) & 60.68~($-$0.26) & 60.69~($-$0.25) & 60.38~($-$0.56) \\
            \cmidrule(l{1pt}r{1pt}){1-1} \cmidrule(l{2pt}r{2pt}){2-2} \cmidrule(l{2pt}r{2pt}){3-3} \cmidrule(l{2pt}r{2pt}){4-4} \cmidrule(l{2pt}r{2pt}){5-5} \cmidrule(l{2pt}r{1pt}){6-6}
            2  & \textbf{61.61} & 60.45~($-$1.16) & 60.36~($-$1.25) & 60.42~($-$1.19) & 60.01~($-$1.60) \\[1pt]
            4  & \textbf{61.28} & 59.99~($-$1.29) & 59.84~($-$1.44) & 59.91~($-$1.37) & 59.35~($-$1.93) \\[1pt]
            8  & \textbf{60.46} & 58.82~($-$1.64) & 58.68~($-$1.78) & 58.75~($-$1.71) & 58.14~($-$2.32) \\[1pt]
            16 & \textbf{58.53} & 55.83~($-$2.70) & 55.62~($-$2.91) & 55.90~($-$2.63) & 55.15~($-$3.38) \\
            \bottomrule
        \end{tabular}
    }
\end{table}
Based on this motivation, we introduce a feature-derived variant \textrm{\textbf{FeatMRL}} for comparison:
\begin{equation}
\mU_{\mathrm{Feat}}^{(t)}
=
\operatorname{\mathbf{Eigen}}\!\left(
\mX^{(t)\top}\mX^{(t)}
\right),
\end{equation}
where the basis $\mU_{\mathrm{Feat}}^{(t)}$ is obtained directly from the Gram matrix of the current input batch $\mX^{(t)}$.
We further consider a momentum-based variant to mitigate potential batch-to-batch variations:
\begin{equation}
\mU_{\mathrm{Feat}}^{(t)} = \operatorname{\mathbf{Eigen}}\,(\mM^{(t)}), \quad \textrm{where} \quad
\mM^{(t)}
=
\beta\,\mM^{(t-1)}
+
(1-\beta)\,\mX^{(t)\top}\mX^{(t)}.
\end{equation}
Table~\ref{tab:featmrl} reports the results, where $\beta=0$ denotes vanilla FeatMRL.
Detailed results for all evaluated $\beta$ values are provided in the Table~\ref{tab:featmrl-breakdown}.
FeatMRL improves over vanilla MRL at several reduced dimensions,
but remains below \name\ at every evaluated dimension.
Momentum smoothing does not close this gap.
These results favor the learned parameter-derived basis over the
feature-derived alternatives evaluated here.

\paragraph{Over-Parameterization of $\mW_{\mathrm{Proj}}$ Brings Additional Gains.} 
In principle, any choice of $R \geq D$ is sufficient for $\mW_{\mathrm{Proj}}$ to attain full row rank.
However, \name\ adopts an over-parameterized design with $R \gg D$.
This is motivated by our observation that increasing $R$ beyond this threshold can further improve the performance of \name.

\begin{wraptable}{r}{0.62\textwidth}
    \vspace{-12pt}
    \centering
    \small
    \caption{Hyperparameter search for the over-parameterization scale of $\mW_{\mathrm{Proj}}$. Darker shades indicate better relative performance; the top-two results are shown in \textbf{bold} and \underline{underline}.}
    \vskip -0.05in
    \label{tab:over-param}
    \setlength{\tabcolsep}{6pt}
    \renewcommand{\arraystretch}{1.08}
    \resizebox{0.92\linewidth}{!}{
        \begin{tabular}{c|cccccc}
        \toprule
        $\tau$ & $\mathrm{D}$ & $\mathrm{4D}$ & $\mathrm{16D}$ & $\mathrm{64D}$ & $\mathrm{128D}$ & $\mathrm{256D}$ \\
        \midrule
        1  & \cellcolor{gridblue!8}60.77 & \cellcolor{gridblue!4}60.73 & \cellcolor{gridblue!16}\underline{60.92} & \cellcolor{gridblue!2}60.72 & \cellcolor{gridblue!16}\textbf{60.94} & \cellcolor{gridblue!14}60.90 \\
        2  & \cellcolor{gridblue!2}61.02 & \cellcolor{gridblue!7}61.27 & \cellcolor{gridblue!10}61.34 & \cellcolor{gridblue!12}61.45 & \cellcolor{gridblue!16}\textbf{61.61} & \cellcolor{gridblue!15}\underline{61.55} \\
        4  & \cellcolor{gridblue!2}60.65 & \cellcolor{gridblue!6}60.84 & \cellcolor{gridblue!10}61.04 & \cellcolor{gridblue!12}61.14 & \cellcolor{gridblue!16}\textbf{61.28} & \cellcolor{gridblue!16}\underline{61.26} \\
        8  & \cellcolor{gridblue!2}59.64 & \cellcolor{gridblue!6}60.00 & \cellcolor{gridblue!10}60.27 & \cellcolor{gridblue!14}60.43 & \cellcolor{gridblue!15}\textbf{60.46} & \cellcolor{gridblue!16}\underline{60.44} \\
        16 & \cellcolor{gridblue!2}57.80 & \cellcolor{gridblue!7}58.12 & \cellcolor{gridblue!10}58.30 & \cellcolor{gridblue!12}58.38 & \cellcolor{gridblue!16}\textbf{58.53} & \cellcolor{gridblue!14}\underline{58.41} \\
        \midrule
        AVG. & \cellcolor{gridblue!2}59.98 & \cellcolor{gridblue!6}60.19 & \cellcolor{gridblue!10}60.37 & \cellcolor{gridblue!11}60.42 & \cellcolor{gridblue!16}\textbf{60.56} & \cellcolor{gridblue!15}\underline{60.51} \\
        \bottomrule
        \end{tabular}
    }
    \vskip -0.14in
\end{wraptable}
To study the effect of $R$, we conduct a hyperparameter search over $R$ on Qwen3-0.6B and report the performance on MTEB.
As shown in Table~\ref{tab:over-param}, we observe that (1) over-parameterizing $\mW_{\mathrm{Proj}}$ improves performance, 
and (2) larger values of $R$ continue to improve performance, while the benefit of further scaling decreases.
Based on these observations, we use $R=128\mathrm{D}$ throughout this paper.

One possible explanation for this advantage is that \name\ makes it feasible to extend MRL to higher-dimensional spaces.
$\mW_{\mathrm{Proj}}$ expands the feature space from $\mathbb{R}^D$ to $\mathbb{R}^R$, while \name\ avoids the need to directly perform MRL at dimension $R$.
This enables finer-grained similarity modeling at a manageable computational cost, thereby extending the model capacity beyond that of traditional MRL.
Similar benefits of over-parameterized designs have also been observed in prior representation learning studies~\citep{simclr}.

\subsubsection{Efficiency and Numerical Stability}
\label{sec:efficiency}
\begin{wraptable}{r}{0.46\textwidth}
    \centering
    \small
    \vskip -0.1in
    \caption{Training throughput degradation of \name\ with different eigen solvers.}
    \label{tab:throughput}
    \resizebox{0.98\linewidth}{!}{
        \begin{tabular}{l|ccc}
            \toprule
            Solver & Qwen3 & Qwen2 & Llama3 \\
            \midrule
            SVD   & $-4.34\%$ & $-3.36\%$ & $-8.35\%$ \\
            {Eigh} & {$-4.59\%$} & {$-3.52\%$} & {$-4.57\%$} \\
            \midrule
            Power & $-0.62\%$ & $-0.53\%$ & $-0.74\%$ \\
            \bottomrule
        \end{tabular}
    }
\end{wraptable}
The over-parameterized design, together with the complexity of eigen decomposition, naturally raises concerns about \name's end-to-end efficiency.
In practice, power iteration confines most of the additional computation to standard matrix operations supported by modern hardware, keeping the overhead affordable.
Moreover, this overhead is input-independent and remains small relative to the backbone computation.
We provide a more comprehensive complexity analysis in Appendix~\ref{app:complexity}.

Table~\ref{tab:throughput} reports \name's training throughput drop relative to MRL for different eigen solvers in the main experiments.
Specifically, we benchmark \texttt{SVD} and \texttt{Eigh} using \texttt{torch.linalg.svd} and \texttt{torch.linalg.eigh}, respectively. For \texttt{Eigh}, we use the MAGMA backend following PyTorch's guidance on numerical stability~\citep{pytorch}.
As shown in Table~\ref{tab:throughput}, our design incurs less overhead than established eigen solvers, achieving throughput nearly identical to that of MRL.
Beyond efficiency, these eigen solvers are also unstable when the eigen gap $|\lambda_i-\lambda_j|$ vanishes.

Formally, for any eigenvector $\vu_i$, its gradient with respect to the Gram matrix takes the form:
\begin{equation}
    \frac{\partial \vu_i}{\partial \mA}[\Delta \mA]
    =
    \sum_{j \neq i}
    \frac{
        \vu_j \vu_j^\top \Delta \mA \vu_i
    }{
        \lambda_i - \lambda_j
    }, \quad \textrm{where} \quad \mA=\mW_{\mathrm{Proj}} \, \mW_{\mathrm{Proj}}^\top.
    \label{eq:eigen-gradient}
\end{equation}
Our implementation avoids this gradient dependence, thereby mitigating sensitivity to the ill-conditioned eigenvalues and improving numerical stability of \name.

At inference time, once the basis is fixed, \name\ incurs only the overhead of a $D \times D$ matrix multiplication.
Meanwhile, it halves the dimension without compromising performance, leading to faster vector retrieval.
Our evaluation confirms that \name\ incurs negligible overhead while providing a retrieval speedup of up to {$1.86\times$}. 
Section~\ref{app:inference-efficiency} provides the detailed experimental setup.

\subsection{Ablation Studies}
We conduct ablation studies to further validate the effectiveness of the design choices behind \name\ and better understand why they lead to improved performance.
\begin{table}[h]
    \caption{Results of ablation studies. Parenthetical values denote drops relative to \name.}
    \label{tab:ablation}
    \centering
    \small
    \resizebox{0.72\textwidth}{!}{
        \begin{tabular}{c|c|c|c|c}
            \toprule
            $\mathbf{\tau}$ & MRL & \name & (a)~MRL \textit{w.} \textrm{Gram} & (b)~MRL \textit{w.} \textrm{Orth} \\
            \midrule
            1  & 60.97 & 60.94 & {61.35~($+$0.41)} & 60.59~($-$0.35) \\
            \cmidrule(l{1pt}r{1pt}){1-1} \cmidrule(l{2pt}r{2pt}){2-3} \cmidrule(l{2pt}r{1pt}){4-5}
            2  & 60.54 & \textbf{61.61} & {60.80~($-$0.81)} & 60.91~($-$0.70) \\
            4  & 59.40 & 61.28 & {59.64~($-$1.64)} & 60.24~($-$1.04) \\
            8  & 57.88 & 60.46 & {57.70~($-$2.76)} & 59.28~($-$1.18) \\
            16 & 54.73 & 58.53 & {54.11~($-$4.42)} & 57.19~($-$1.34) \\
            \bottomrule
        \end{tabular}
    }
    \vspace{-8pt}
\end{table}
\subsubsection{Spectrum Contributes Beyond Gram-matrix Over-Parameterization.}
Compared with vanilla MRL, \name\ introduces a learnable factor
$\mW_{\mathrm{Proj}}\in\mathbb{R}^{D\times R}$ and derives its
ordered spectral basis from the corresponding Gram matrix.
To examine whether the improvement can be explained merely by this
additional parameterization, we construct an ablated variant that
retains the same learnable factor but removes spectral decomposition.

For a nested dimension $m$, let
$\mW_m=\mW_{\mathrm{Proj}}[:m,:]$.
This variant retains the original coordinate prefixes $\mX^{(m)}$
and evaluates similarity after the implicit lifting
$\mX^{(m)}\mW_m$.
Equivalently, its similarity is induced by the Gram matrix
$\mG_m=\mW_m\mW_m^\top$.
We therefore refer to this variant as MRL \textit{w.} Gram.
The full formulation and details are provided in
Appendix~\ref{app:implicit-gram}.

As shown in Table~\ref{tab:ablation}, the Gram-based variant provides
only limited improvements over vanilla MRL and degrades rapidly under
stronger compression.
In contrast, \name\ consistently performs better at reduced dimensions.
These results show that introducing the same learnable factor through Gram-based similarity does not reproduce the gains of \name, supporting the proposed spectral readout over this alternative. 
Extra comparison with a learnable linear is provided in Appendix~\ref{app:linear-readout}.

\subsubsection{Eigen Decomposition Provides More Than Learnable Orthogonality}
Eigen decomposition organizes the learned structure into an ordered set of orthogonal directions.
This suggests two possible sources of improvement: learning a favorable orthogonal basis and imposing a spectral ordering.
To disentangle their effects, we construct an ablated variant that retains only learnable orthogonality.
This allows us to assess whether the learned spectral ordering benefits MRL's nested readouts.
Specifically, we replace the orthogonal basis in Eq.~\ref{eq:eigh} as:
\begin{equation}
    \mU_{\mathrm{Orth}} \, = \, \operatorname{\mathbf{QR}}\!\left(\,\mW_{\mathrm{Proj}} \, \mW_{\mathrm{Proj}}^\top\,\right),
\end{equation}
where QR is applied to the Gram matrix alone to retain orthogonality. Table~\ref{tab:ablation}~(b) reports its results.
Compared with vanilla MRL, learned orthogonality provides a clear advantage, but remains inferior to the full \name\ design.
This further highlights the value of ordering feature directions by learned importance, which is naturally compatible with MRL's nested readout design.

\subsection{Validate \name\ at Larger Dimensions and Training Scales}
\label{sec:underperform}
We scale both model width and training scale to further evaluate \name\ from two perspectives: (1) whether its effectiveness persists in larger representation spaces, and (2) whether its gains truly extend the performance frontier of MRL rather than merely resulting from faster convergence.

\begin{figure}[t]
    \vspace{-8pt}
    \centering
    \begin{minipage}{0.46\linewidth}
        \centering
        \captionof{table}{Comparison between MRL and \name\ on Qwen3-4B. \name\ remains effective as the representation space scales up.}
        \label{tab:4b}
        \resizebox{0.92\linewidth}{!}{
            \begin{tabular}{c|c|c|c}
                \toprule
                \multicolumn{4}{c}{\textbf{\textrm{Qwen3-4B}}} \\
                \midrule
                $\mathbf{D}$ & \textbf{Bands} & MRL & \name \\
                \midrule
                2560 & $0 : 20$ & 64.71 & 64.29~($-$0.42) \\
                \cmidrule(l{1pt}r{1pt}){1-1} 
                \cmidrule(l{2pt}r{2pt}){2-2} 
                \cmidrule(l{2pt}r{1pt}){3-4}
                2432 & $0 : 19$ & 64.73 & \cellcolor{gray!12}64.98~($+$0.25) \\
                \cmidrule(l{1pt}r{1pt}){1-1} 
                \cmidrule(l{2pt}r{2pt}){2-2} 
                \cmidrule(l{2pt}r{1pt}){3-4}
                1024 & $0 : \,8$ & 64.38 & \cellcolor{gray!12}\textbf{65.06}~(+0.68) \\[1pt]
                512  & $0 : \,4$ & 63.51 & \cellcolor{gray!12}64.86~(+1.35) \\[1pt]
                256  & $0 : \,2$ & 62.40 & \cellcolor{gray!12}64.49~(+2.09) \\[1pt]
                128  & $0 : \,1$ & 60.20 & \cellcolor{gray!12}63.51~(+3.31) \\
                \bottomrule
            \end{tabular}
        }
    \end{minipage}
    \hfill
    \begin{minipage}{0.5\linewidth}
        \centering
        \includegraphics[width=\linewidth]{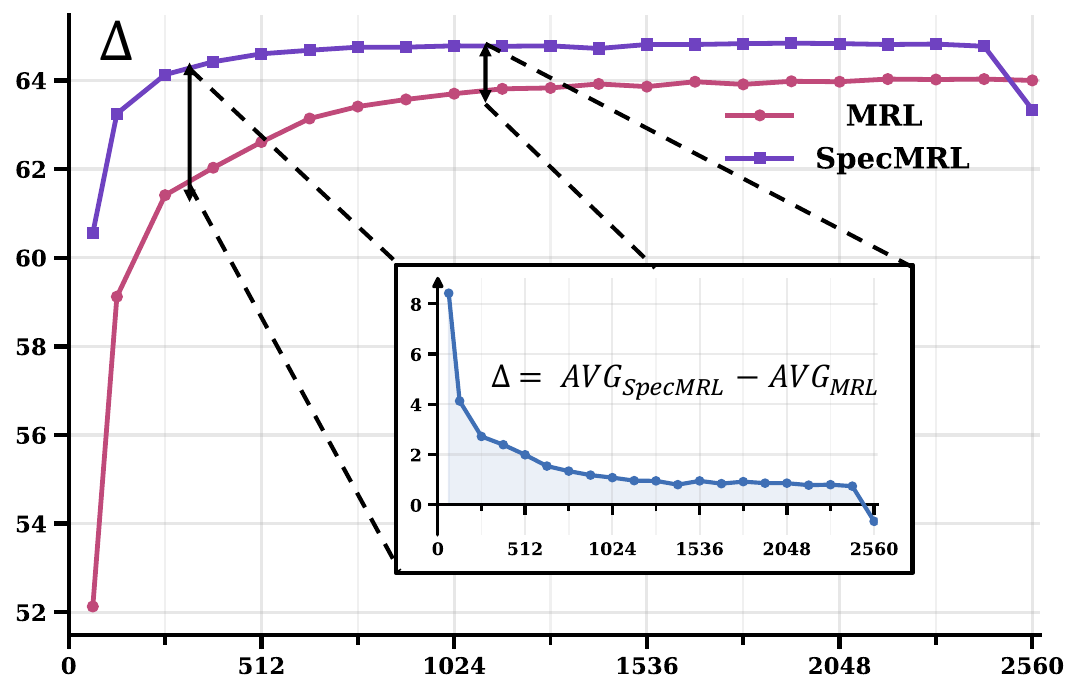}
        \captionof{figure}{MTEB performance trends of \name\ across nested dimensions. The final spectral band introduces an unexpected performance drop.}
        \label{fig:alldim}
    \end{minipage}
    \vspace{-12pt}
\end{figure}


\paragraph{Empirical Observations at Larger Representation Dimensions.}
\label{sec:scalabilty}
We first increase the model width and scale the Qwen3 backbone from 0.6B to 4B to extend our evaluation.
Table~\ref{tab:4b} compares \name\ with MRL on Qwen3-4B using MTEB.
The relative gains of \name\ over MRL follow trends similar to those observed on the 0.6B model in the main results reported in Table~\ref{tab:main}.

Leveraging the larger hidden size of Qwen3-4B, we further conduct a fine-grained sweep over nested dimensions to investigate \name's behavior under different spectral truncations.
Specifically, we partition the spectral basis into 20 bands and report the average MTEB score over a subset of tasks with the first \(n\) bands retained.
Figure~\ref{fig:alldim} visualizes the resulting trend, where \name\ exhibits a pattern similar to that of MRL: enlarging the representation leads to improved performance for most retained spectral bands.
Moreover, the learned spectral basis remains effective across unseen dimensions, indicating stability at dimensions unseen during training.

However, we observe an unexpected degradation when the final spectral bands are retained.
We therefore remove these bands in other experimental settings to examine whether the same effect persists.
Table~\ref{tab:filter} summarizes the results, where \(i:j\) denotes retaining bands \(i\) through \(j-1\).
\begin{table}[h]
    \caption{Comparison of \name\ performance with and without the final spectral bands.}
    \label{tab:filter}
    \centering
    \small
    \resizebox{0.86\textwidth}{!}{
        \begin{tabular}{c|c|c|c|c|c|c}
            \toprule
            & \multicolumn{2}{c}{\textbf{\textrm{Qwen3-0.6B}}} & \multicolumn{2}{c}{\textbf{\textrm{Qwen2-0.5B}}} & \multicolumn{2}{c}{\textbf{\textrm{Llama3.2-1B}}} \\
            \cmidrule(l{1pt}r{1pt}){1-1} \cmidrule(l{2pt}r{2pt}){2-3} \cmidrule(l{2pt}r{2pt}){4-5} \cmidrule(l{2pt}r{1pt}){6-7}
            $\mathrm{Bands}$ & MRL & \name & MRL & \name & MRL & \name \\
            \midrule
            $0 : 20$ & 60.97 & 60.94~($-$0.03) & 58.96 & 58.74~($-$0.22) & 61.92 & 61.87~($-$0.05) \\
            \cmidrule(l{1pt}r{1pt}){1-1} \cmidrule(l{2pt}r{2pt}){2-3} \cmidrule(l{2pt}r{2pt}){4-5} \cmidrule(l{2pt}r{1pt}){6-7}
            $0 : 19$ & {60.94} & {61.35~($+$0.41)} & {59.00} & {59.20~($+$0.20)} & {61.93} & {62.37~($+$0.44)} \\
            \cmidrule(l{1pt}r{1pt}){1-1} \cmidrule(l{2pt}r{2pt}){2-3} \cmidrule(l{2pt}r{2pt}){4-5} \cmidrule(l{2pt}r{1pt}){6-7}
            $0 : 10$ & 60.54 & \cellcolor{gray!12}\textbf{61.61}~(+1.07) & 58.67 & \cellcolor{gray!12}\textbf{59.44}~(+0.77) & 61.74 & \cellcolor{gray!12}\textbf{62.42}~(+0.68) \\
            \bottomrule
        \end{tabular}
    }
\end{table}

\begin{wrapfigure}{r}{0.46\textwidth}
    \vspace{-12pt}
    \centering
    \includegraphics[width=0.98\linewidth]{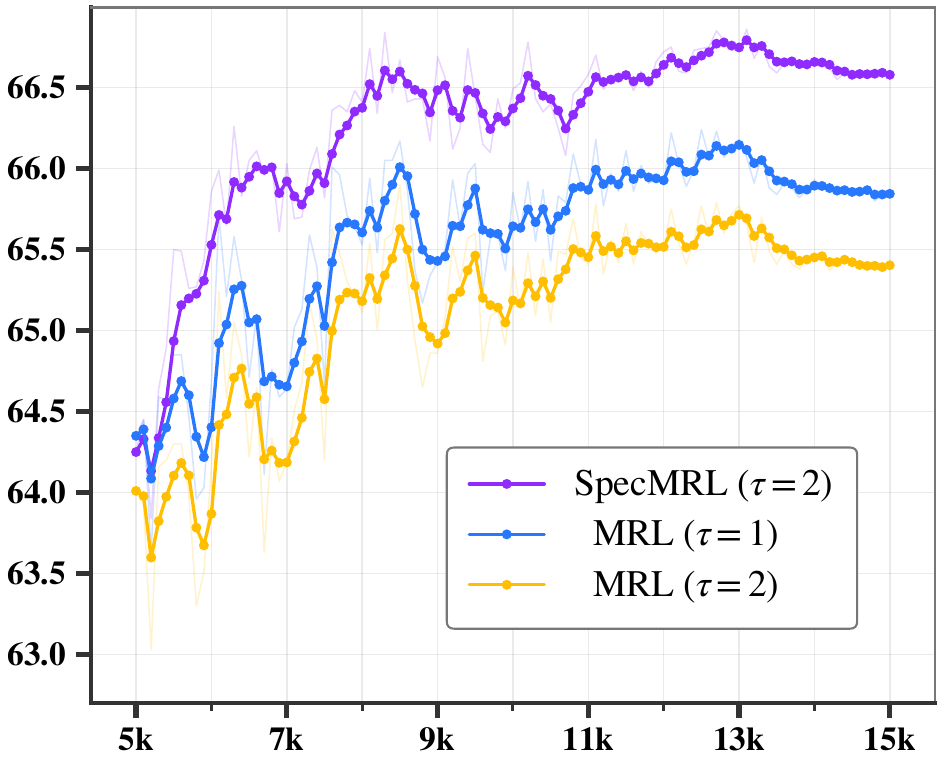}
    \vspace{-4pt}
    \caption{MTEB subset results evaluated every 100 steps. The advantage of \name\ is consistent throughout the training process. }
    \label{fig:allstep}
    \vspace{-22pt}
\end{wrapfigure}

The degradation in the smallest-singular-value bands remains consistent across different settings.
Similar behavior was observed for the spectrum of the LLM unembedding matrix~\citep{embedfilter}.
Our further analysis attributes this phenomenon to disproportionate projection onto the corresponding spectral subspace, which partially disrupts isotropy.
We defer this analysis to Appendix~\ref{app:projection-energy} and hope these findings may motivate more principled designs for embedding training.

\paragraph{\name\ Extends the Performance Frontier Beyond Faster Convergence.}
We next investigate whether \name's gains persist beyond faster convergence at larger training scales.
Therefore, we extend Qwen3-0.6B training beyond main experiments and monitor its downstream performance relative to MRL in an over-training regime.
As shown in Figure~\ref{fig:allstep}, \name\ not only retains its advantage over MRL throughout training but also attains a higher performance ceiling.
Table~\ref{tab:allstep-breakdown} reports the full MTEB evaluation in an over-training regime.

Taken together, these results highlight the potential of \name\ toward training with larger models and longer horizons.
We hope to further validate this scalability under larger computational budgets.

\subsection{Comparison of \name\ with Existing Improvements to MRL}
\label{sec:ese}
\begin{table}[h]
    \caption{Comparison of \name\ with MRL variants on Qwen3-0.6B.}
    \label{tab:mrl-improvements}
    \centering
    \small
    \resizebox{0.68\textwidth}{!}{
        \begin{tabular}{c|c|c|c|c}
            \toprule
            $\boldsymbol{\tau}$ & MRL & ESE (L2C) & MIPIC (SIA) & \name \\
            \midrule
            1  & \textbf{60.97}
               & 57.97~($-$3.00)
               & 60.42~($-$0.55)
               & 60.94~($-$0.03) \\
            \cmidrule(l{1pt}r{1pt}){1-1}
            \cmidrule(l{2pt}r{1pt}){2-5}
            2  & 60.54
               & 57.26~($-$3.28)
               & 59.93~($-$0.61)
               & \cellcolor{gray!12}\textbf{61.61}~($+$1.07) \\[1pt]
            4  & 59.40
               & 56.38~($-$3.02)
               & 59.08~($-$0.32)
               & \cellcolor{gray!12}\textbf{61.28}~($+$1.88) \\[1pt]
            8  & 57.88
               & 55.18~($-$2.70)
               & 57.74~($-$0.14)
               & \cellcolor{gray!12}\textbf{60.46}~($+$2.58) \\[1pt]
            16 & 54.73
               & 50.26~($-$4.47)
               & 55.09~($+$0.36)
               & \cellcolor{gray!12}\textbf{58.53}~($+$3.80) \\
            \bottomrule
        \end{tabular}
    }
\end{table}
We also compare \name\ with other MRL variants (ESE, \citealp{ese}; MIPIC, \citealp{mipic}) to further validate its effectiveness.
Both ESE and MIPIC are designed to enhance information retention at lower dimensions through additional supervision.
Specifically, they augment the standard MRL objective with different loss designs for cross-dimensional supervision.
Given that, we restrict our comparison to their reduced-dimension variants, ESE (L2C) and MIPIC (SIA), excluding other design elements.
Further discussion of their full designs is provided in Section~\ref{sec:related}.

We compare \name\ with these baselines on Qwen3-0.6B using the same setup as in our main experiments.
Table~\ref{tab:mrl-improvements} presents the comparative results, where \name\ achieves the best overall performance among the methods considered.
Notably, ESE (L2C) shows a performance collapse on classification tasks.
Therefore, we also report the average over the non-classification subset in Table~\ref{tab:mrl-improvements-breakdown}, where the same conclusion holds.
Beyond these gains, \name\ provides additional advantage in practice, since it only changes the prefix-readout basis.
As a consequence, this makes \name\ a compatible extension to MRL, requiring no extra hyperparameter tuning while remaining robust across diverse tasks.
Further details for baselines implementation and additional evaluation results are provided in Appendix~\ref{app:mrl-baselines} for clarity.



\subsection{Robustness to MRL Configurations}
\label{sec:mrl-setup}


We further evaluate \name\ under different MRL settings to assess whether its advantages are robust across configurations.
Specifically, we conduct two experiments on Qwen3-0.6B: (1) setting all nested loss weights to $1$, and (2) adopting sparse supervision by dropping the last two dimensions ($128$ and $64$) from the nested training set.
All other training settings are kept unchanged.

Table~\ref{tab:mrl-config} compares MRL and \name\ under different configurations. Across these configurations, \name\ consistently preserves the low-dimensional advantage observed in the main experiments, indicating robustness to MRL training settings.
Notably, under sparse supervision, SpecMRL achieves larger gains over MRL at the two prefixes without dedicated loss terms, including a $7.29$-point improvement at $64$ dimensions.
This suggests that \name\ depends less on direct supervision at individual prefixes to preserve low-dimensional information, enabling effective readout under sparser nested supervision and demonstrating robustness to the choice of training dimensions.

\begin{table}[htbp]
    \caption{
        Robustness to different MRL training configurations.
    }
    \label{tab:mrl-config}
    \centering
    \small
    \resizebox{0.88\textwidth}{!}{
        \begin{tabular}{c|cc|cc|cc}
            \toprule
            & \multicolumn{2}{c|}{\textbf{\textrm{Default}}}
            & \multicolumn{2}{c|}{\textbf{\textrm{Uniform Weights}}}
            & \multicolumn{2}{c}{\textbf{\textrm{Sparse Supervision}}} \\
            \cmidrule(l{1pt}r{1pt}){1-1}
            \cmidrule(l{2pt}r{2pt}){2-3}
            \cmidrule(l{2pt}r{2pt}){4-5}
            \cmidrule(l{2pt}r{1pt}){6-7}

            $\boldsymbol{\tau}$
            & MRL & \name
            & MRL & \name
            & MRL & \name \\
            \midrule

            1
            & 60.97 & 60.94~($-$0.03)
            & 60.35 & 60.37~(+0.02)
            & 60.73 & 60.76~($+$0.03) \\

            \cmidrule(l{1pt}r{1pt}){1-1}
            \cmidrule(l{2pt}r{2pt}){2-3}
            \cmidrule(l{2pt}r{2pt}){4-5}
            \cmidrule(l{2pt}r{1pt}){6-7}

            2
            & 60.54 & \cellcolor{gray!12}\textbf{61.61}~(+1.07)
            & 60.09 & \cellcolor{gray!12}\textbf{61.79}~(+1.70)
            & 60.11 & \cellcolor{gray!12}\textbf{61.53}~(+1.42) \\[1pt]

            4
            & 59.40 & \cellcolor{gray!12}61.28~(+1.88)
            & 59.31 & \cellcolor{gray!12}61.60~(+2.29)
            & 58.76 & \cellcolor{gray!12}61.20~(+2.44) \\[1pt]

            8
            & 57.88 & \cellcolor{gray!12}60.46~(+2.58)
            & 58.38 & \cellcolor{gray!12}61.04~(+2.66)
            & 55.88 & \cellcolor{gray!12}60.15~(+4.27) \\[1pt]

            16
            & 54.73 & \cellcolor{gray!12}58.53~(+3.80)
            & 56.48 & \cellcolor{gray!12}59.30~(+2.82)
            & 49.94 & \cellcolor{gray!12}57.23~(+7.29) \\

            \bottomrule
        \end{tabular}
    }
\end{table}


\subsection{Inference and Retrieval Efficiency}
\label{app:inference-efficiency}

\paragraph{Embedding Encoding.}
We compare MRL and \name\ with Qwen3-0.6B on the same 100,000 passages.
The measured time includes the forward pass and the additional projection in \name.
With a precomputed spectral basis, \name\ performs a full
$1024\times1024$ projection yet increases total encoding time
from $312.85$\,s for MRL to $313.99$\,s, corresponding to an
overhead of only $0.36\%$.

\paragraph{Vector Retrieval.}
We randomly select 8,192 passages from one million MS MARCO passages as queries and use the remainder as the retrieval database.
Using FAISS's \texttt{GpuIndexFlatIP} on a single NVIDIA A100 PCIe 40GB GPU, we perform FP32 exact top-100 retrieval.
Indices are prebuilt and remain in GPU memory.
We measure retrieval throughput in queries per second (QPS) across different dimensions and query batch sizes.
As shown in Table~\ref{tab:retrieval-throughput}, speedups generally increase with query batch size before leveling off.
Using half the dimensions, \name\ achieves the best score (Table~\ref{tab:main}) and up to a $1.86\times$ retrieval speedup, improving both quality and efficiency.

\begin{table*}[htbp]
    \centering
    \caption{Retrieval throughput (QPS) across compression ratios $\tau$
    and query batch sizes $B$, with output dimension $1024/\tau$.
    Parentheses show speedups over $\tau=1$ at the same $B$.
    Bold highlights the retrieval speedup of the representation with the highest average MTEB score in Table~\ref{tab:main}.
    }
    \label{tab:retrieval-throughput}
    \small
    \setlength{\tabcolsep}{6pt}
    \renewcommand{\arraystretch}{1.15}
    \begin{tabular}{crrrrr}
        \toprule
        $\tau$
        & \multicolumn{1}{c}{$B=1$}
        & \multicolumn{1}{c}{$B=64$}
        & \multicolumn{1}{c}{$B=256$}
        & \multicolumn{1}{c}{$B=1024$}
        & \multicolumn{1}{c}{$B=2048$} \\
        \midrule
        1
        & 140~($1.00\times$)
        & 5,207~($1.00\times$)
        & 7,103~($1.00\times$)
        & 7,859~($1.00\times$)
        & 7,961~($1.00\times$) \\
        \midrule
        2
        & 176~($1.26\times$)
        & 7,713~($1.48\times$)
        & 12,558~($1.77\times$)
        & 14,605~($1.86\times$)
        & \cellcolor{gray!16}14,791~($\boldsymbol{1.86\times}$) \\

        4
        & 205~($1.46\times$)
        & 9,891~($1.90\times$)
        & 20,174~($2.84\times$)
        & 25,508~($3.25\times$)
        & \cellcolor{gray!8}25,906~(${3.25\times}$) \\

        8
        & 231~($1.65\times$)
        & 11,414~($2.19\times$)
        & 28,986~($4.08\times$)
        & 40,594~($5.17\times$)
        & \cellcolor{gray!8}41,282~($5.19\times$) \\

        16
        & 352~($2.52\times$)
        & 12,319~($2.37\times$)
        & 36,712~($5.17\times$)
        & 56,884~($7.24\times$)
        & \cellcolor{gray!8}57,814~($7.26\times$) \\
        \bottomrule
    \end{tabular}
    \vspace{-8pt}
\end{table*}

\section{Related Work}
\label{sec:related}


Modern applications use models to encode data into embeddings for a wide range of downstream tasks.
Despite continued improvements in quality, the storage and computational costs of fixed, high-dimensional embeddings remain a practical bottleneck. MRL~\citep{mrl} addresses this limitation through joint supervision of nested prefixes, enabling flexible output dimensions.
While MRL is simple, several methods seek further improvements by introducing auxiliary objectives.
ESE~\citep{ese} introduces a learn-to-compress (L2C) objective that supervises truncated prefixes with PCA-based compression targets. 
Similarly, MIPIC~\citep{mipic} aligns token-level relations across dimensions through self-distilled intra-relational alignment (SIA), matching token-importance distributions with KL divergence and inter-token geometry with centered kernel alignment. 
Both methods derive alignment targets from full-dimensional embeddings to supervise low-dimensional prefixes, while also incorporating intermediate-layer supervision to improve embeddings at shallower depths.
In contrast to alignment-based supervision, MIC~\citep{mic} introduces geometric regularization to reduce redundancy between prefix and residual dimensions.


MRL has also been extended to frozen models by adding lightweight trainable layers, supporting effective prefix truncation at multiple dimensions without backbone update.
Following this idea, 
Matryoshka-Adaptor~\citep{mrl-adaptor} learns nested embeddings through lightweight adapters, while SMEC~\citep{smec} further improves this adaptation process through sequential training and adaptive dimension selection.
More recently, AdaWidth~\citep{adawidth} learns an orthogonal rotation of frozen embeddings, providing a learnable orthogonal basis without explicit spectral ordering. This design is conceptually closely related to the learnable-basis ablation considered in our analysis, which underperforms the proposed spectral basis.

\section{Conclusion}


We present SpecMRL, a parametric-spectral redesign of MRL that addresses the limitations of fixed basis-prefix readout.
By exploiting the ordered orthogonal structure of the learned spectrum, SpecMRL reorganizes representations into a form naturally suited to nested truncation.
Across diverse experimental settings, SpecMRL consistently outperforms MRL at reduced dimensions with negligible overhead, while further extending the performance frontier.
These findings highlight the readout space as an important learning target in nested representation. We hope this perspective motivates further exploration of information organization and inspires more principled MRL designs.


\bibliography{iclr2027_conference}
\bibliographystyle{iclr2027_conference}

\appendix



\newpage

\section{Implementation Details}
\label{app:setup}

\subsection{General Setup}
\paragraph{Training Setup.}
We conduct the main experiments reported in Table~\ref{tab:main} using backbones from widely used model families, including Qwen3-0.6B-Base~\citep{qwen3}, Qwen2-0.5B~\citep{qwen2}, and Llama-3.2-1B~\citep{llama3}. 
We further include Qwen3-4B-Base in the scaling experiments to evaluate performance at a larger model scale.
Unless otherwise specified,  we instantiate $\mathcal{\mathbf{P}}$ as mean pooling to derive the embeddings. 
For MRL training setup, we follow the setup of~\citet{kalm} and summarize the configuration for each backbone in Table~\ref{tab:mrl_setup}, where $m_i$ denotes the $i$-th nested dimension and $w_i$ denotes its corresponding loss weight.
For brevity, $\tau$ denotes the compression ratio rounded to the nearest power of two throughout the tables.

We provide further details on our training setup.
For data preparation, we use the publicly available KaLM-embedding-finetuning-data dataset\footnote{\url{huggingface.co/datasets/KaLM-Embedding/KaLM-embedding-finetuning-data}}~\citep{kalm}.
By default, we train on a randomly sampled subset of 1,152,000 examples, {approximately $1/5$ of the full dataset}, to keep the experimental cost manageable.
Specifically, each training example consists of one query, one positive example, and seven hard negatives.
We also use in-batch negatives across GPUs.
For the long-run experiments in Section~\ref{sec:scalabilty}, we use the full dataset to evaluate performance under extended training.

Regarding the training hyperparameters, the per-device batch size is set to 96 queries, resulting in a global batch size of 384 across four GPUs. 
We adopt GradCache~\citep{gradcache} to reduce memory usage.
The maximum sequence length is set to 512 tokens for both queries and passages.
We use the AdamW optimizer~\citep{adam,adamw} with a learning rate of $10^{-4}$, a warmup ratio of 0.1 followed by linear decay, and a weight decay of 0.01. 
For contrastive learning, the temperature is set to 0.01 across all experimental settings.
We initialize the entries of $\mW_{\mathrm{Proj}}$ from $\mathcal{N}(0,0.02^2)$ and obtain the initial spectral basis via the eigenvectors of its Gram matrix.
More implementation details and reproducibility information are provided in our codebase.


\paragraph{Evaluation.}
We conduct experiments on the widely used English MTEB benchmark~\citep{mteb}, which comprises 56 datasets spanning seven task categories: Semantic Textual Similarity~(\textbf{STS.}), Classification~(\textbf{Class.}), Clustering~(\textbf{Cluster.}), Pair Classification~(\textbf{PairClass.}), Reranking~(\textbf{Rerank.}), Retrieval~(\textbf{Retr.}), and Summarization~(\textbf{Sum.}).
For experiments presented in Figures~\ref{fig:alldim} and~\ref{fig:allstep}, we report the average score over a 15-dataset evaluation subset detailed in Table~\ref{tab:mteb_setup}.

For sentence encoding, both MRL and \name\ follow KaLM's task-specific inference instructions~\citep{kalm}.
For \name, we perform one orthogonal iteration in FP32 using the final parameter Gram matrix, starting from the spectral basis saved in the checkpoint.
The resulting basis remains fixed throughout evaluation.
For Qwen2-0.5B, we use the same nested dimensions for training and evaluation, following KaLM~\citep{kalm}.
We use unnormalized embeddings for classification and apply $L_2$ normalization at each target dimension for all other tasks.

\subsection{Implementation of ESE and MIPIC}
\label{app:mrl-baselines}
To focus on dimensional compression and facilitate a controlled comparison with MRL and \name, we adapt ESE and MIPIC to use only the final-layer representations of Qwen3-0.6B.
Method-specific hyperparameters are selected following the original papers~\citep{ese,mipic}.
The implementations of all baselines are also available in our code repository.
Detailed evaluation results for all compared methods are presented in Table~\ref{tab:mrl-improvements-breakdown}.

\paragraph{ESE (L2C).}
We retain the learn-to-compress objective and apply it to
the final-layer sentence embeddings, omitting intermediate-layer
supervision.
Contrastive losses are applied to the full 1024-dimensional representation and its 128-dimensional prefix.
The latter additionally receives alignment supervision from
the compression targets.
Representations at other dimensions are obtained
by prefix truncation at evaluation time, without additional
training losses at those dimensions.

\vspace{-4pt}

\paragraph{MIPIC (SIA).}
We combine the MRL objective with SIA applied to final-layer token representations.
The nested dimensions and MRL loss weights follow Table~\ref{tab:mrl_setup}.
Using mean-pooled sentence embeddings as anchors, SIA aligns token-importance distributions across dimensions through KL divergence.
Linear CKA further aligns the relational structures of full-dimensional and truncated representations on token subsets selected by full-dimensional importance scores.



\subsection{Implementation of the Gram-Matrix Ablation}
\label{app:implicit-gram}
In this ablated variant, we retain the same learnable projection factor
$\mW_{\mathrm{Proj}}$ as in \name.
For each nested dimension $m$, we define
$\mW_m=\mW_{\mathrm{Proj}}[\,:m,\,:\,]$ and
$\mG_m=\mW_m\,\mW_m^\top$, and let
$\mX^{(m)},\;\mY^{(m)}\in\mathbb{R}^{1\times m}$
denote the corresponding truncated backbone embeddings.
To investigate whether the performance gains of \name\ arise solely
from the introduction of additional learnable parameters, we directly apply
$\mW_m$ to the corresponding truncated backbone embeddings,
without performing any spectral decomposition.
This produces the resulting $R$-dimensional representations
$\mX^{(m)}\,\mW_m$ and $\mY^{(m)}\,\mW_m$,
whose cosine similarity can be computed implicitly as:
\begin{equation}
\operatorname{sim}(\mX^{(m)}\mW_m,\mY^{(m)}\mW_m)
=
\frac{\mX^{(m)}\mG_m(\mY^{(m)})^\top}
{\sqrt{\mX^{(m)}\mG_m(\mX^{(m)})^\top}
 \sqrt{\mY^{(m)}\mG_m(\mY^{(m)})^\top}},
\end{equation}

At evaluation, we compute the Cholesky factorization
$\mG_m=\mL_m\mL_m^\top$
for each target dimension.
Each truncated embedding  $\mX^{(m)}$ is mapped to
the $m$-dimensional embedding $\mX^{(m)}\mL_m$.
This preserves the inner products and norms of the lifted
representations, and hence their cosine similarities.

\section{Comparison with a Learnable Linear Readout}
\label{app:linear-readout}


\begin{wraptable}{r}{0.46\textwidth}
    \centering
    \small
    \vspace{-8pt}
    \caption{Comparison with learnable readout.}
    \label{tab:linear-readout}
    \resizebox{0.88\linewidth}{!}{
        \begin{tabular}{c|c|c}
            \toprule
            $\mathbf{\tau}$ & \name & MRL \textit{w.} Linear \\
            \midrule
            1  & 60.94
               & 60.92~($-$0.02) \\
            \cmidrule(l{1pt}r{1pt}){1-1}
            \cmidrule(l{2pt}r{1pt}){2-3}
            2  & \cellcolor{gray!12}\textbf{61.61}
               & 60.72~($-$0.89) \\
            4  & \cellcolor{gray!12}61.28
               & 60.14~($-$1.14) \\
            8  & \cellcolor{gray!12}60.46
               & 59.13~($-$1.33) \\
            16 & \cellcolor{gray!12}58.53
               & 57.03~($-$1.50) \\
            \bottomrule
        \end{tabular}
    }
    \vspace{-8pt}
\end{wraptable}

To assess whether a generic learnable readout can reproduce the gains of \name, we augment vanilla MRL with a $D\times D$ linear layer as the readout.
We jointly train this layer with the backbone and denote the resulting as MRL \textit{w.} Linear.

As shown in Table~\ref{tab:linear-readout}, although both readouts can integrate all original channels, MRL \textit{w.} Linear underperforms \name\ at every evaluated dimension, with the gap widening as the dimension decreases. 
In principle, an unconstrained linear layer can represent the orthogonal transformation used by \name.
However, under the same MRL objective, this expressive capacity does not translate into comparable prefix quality, further supporting the value of \name's ordered spectral basis.


\section{Parameter Space Reflects Embedding Geometry}
\label{app:parameter-geometry} 
From a parametric perspective, each column of $\mW_{\mathrm{Proj}}$ can be interpreted as an embedding-space vector, making the weight matrix structurally analogous to a cached collection of embeddings.
We contend that $\mW_{\mathrm{Proj}}$ encodes high-level semantic information, such that their collective organization in parameter space reflects the geometry of the learned embedding space during training.

We present our empirical observations to substantiate this structural correspondence. 
Specifically, we randomly sample $N=10^6$ sentences from the training set and encode them separately using MRL and \name\ to obtain calibration embeddings $\hat{\mathcal{E}}_{\mathrm{cal}} \in \mathbb{R}^{N \times D}$.
Using these embeddings as a reference, we extract their principal directions $\mU_{\hat{\mathcal{E}}}$ and compare the resulting leading $k$-dimensional subspace with the corresponding readout subspace using principal-angle overlap:
\begin{equation}
    S_k
    =
    \frac{1}{k}
    \left\|
        \left[
            \left(\mU_{\hat{\mathcal{E}}}\right)_{:,1:k}
        \right]^\top
        \mU_{:,1:k}
    \right\|_F^2,
    \label{eq:readout-subspace-overlap}
\end{equation}
where $\mU$ denotes the readout basis of the model.
The score lies in $[0,1]$, with higher values indicating greater alignment.
For random $k$-dimensional subspaces in $\mathbb{R}^{D}$, the expected overlap is $k/D$.

\begin{wraptable}{r}{0.46\textwidth}
    \centering
    \small
    \caption{Readout subspace alignment.}
    \label{tab:readout-subspace-overlap}
    \setlength{\tabcolsep}{4pt}
    \renewcommand{\arraystretch}{1.08}
    \resizebox{0.98\linewidth}{!}{
        \begin{tabular}{c|ccc}
            \toprule
            $k$ & Rand & MRL & \name \\
            \midrule
            64  & 0.0628 & 0.2522 & \textbf{0.3025}~(+~5.03\%) \\[1pt]
            128 & 0.1237 & 0.2783 & \textbf{0.4295}~(+~15.12\%) \\[1pt]
            256 & 0.2513 & 0.3171 & \textbf{0.4715}~(+~15.44\%) \\[1pt]
            512 & 0.4999 & 0.5197 & \textbf{0.5518}~(+~3.21\%) \\
            \bottomrule
        \end{tabular}
    }
\end{wraptable}


As shown in Table~\ref{tab:readout-subspace-overlap}, 
both the overlap scores of MRL and \name\ exceed the random baseline, suggesting that the MRL objective encourages alignment between readout subspaces and embedding geometry.
Viewed as a collection of embedding vectors, the $D\times R$ parameter matrix can be interpreted as a learned proxy for the full population of sentence embeddings, fitting their geometry during training.
Compared with the fixed coordinate basis used by MRL,
the spectral readout subspaces of \name\ exhibit stronger alignment with the dominant subspaces of sentence embeddings.
This closer alignment may help preserve informative embedding structure under dimensional readout, thereby partially explaining the improved low-dimensional performance of \name.



For \name, the readout basis used is obtained by power iteration, following the procedure described in Appendix~\ref{app:setup}.
We further verify this approximation against the ordered eigenbasis computed directly from the same Gram matrix.
At the final checkpoint, their subspace overlap exceeds $99.9\%$ for all evaluated $k$, indicating that the two bases span nearly identical leading subspaces.

\section{Energy Concentration in the Tailing Spectral Subspace}
\label{app:projection-energy}
We conduct further analyses to investigate why dropping the final spectral bands can improve \name{}'s downstream performance.
Motivated by the analysis in~\citep{embedfilter}, we examine the behavior of \name{}-derived representations when projected onto different spectral bands.

Specifically, we first partition the spectrum into 20 consecutive bands ordered by eigenvalue.
We then sample $N=10^6$ passages from the MS MARCO corpus~\citep{msmarco} to obtain the calibration embeddings.
Let $z_{ij}=\mathbf{u}_j^\top\mX_{[i]}$ denote the projection
of $\mathbf{X}_{[i]}$ onto the $j$-th spectral direction.
The average energy fraction of band $B_b$ over the $N$ calibration embeddings is:
\begin{equation}
  E_b =
  \frac{1}{N}\sum_{i=1}^{N}
  \frac{\sum_{j\in B_b}z_{ij}^{2}}
       {\sum_{j=1}^{1024}z_{ij}^{2}}.
\end{equation}

\begin{wrapfigure}{r}{0.46\textwidth}
\centering
\vspace{-12pt}
\includegraphics[width=\linewidth]
    {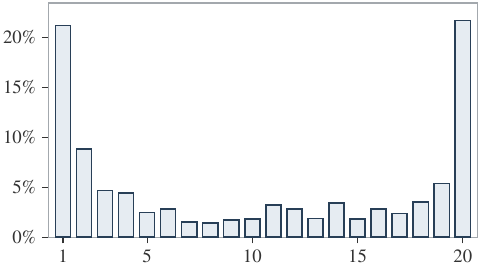}
    \vspace{-20pt}
\caption{Average projection energy fractions.}
\label{fig:projection-energy}
\vspace{-12pt}
\end{wrapfigure}

Figure~\ref{fig:projection-energy} shows that the final band accounts for a large fraction of energy (21.72\%), despite spanning only 5\% of the spectral directions.
Ideally, the energy should decay smoothly across the ordered spectral bands, with the first band accounting for the largest share, consistent with its intended role in capturing the most informative semantic directions in MRL.
The decay pattern reflects the decreasing semantic importance of later spectral bands, while the smoothness helps maintain isotropy by avoiding concentration in a small subset of directions.
In turn, the disproportionate concentration in the final band indicates increased anisotropy, suggesting a departure from the expected geometry.
Similar observations have also been reported by \citet{embedfilter}.
Although the cause of this phenomenon remains unclear, we hope this analysis provides an initial explanation for the observed improvement and motivates further investigation.

\section{Computational Complexity}
\label{app:complexity}
This section provides a supplementary analysis of \name's computational overhead relative to MRL.
The additional computations consist of Gram matrix construction and orthogonal basis update, followed by projecting a batch of $B_{\mathrm{proj}}$ embeddings of dimension $D$.
As shown in Fig.~\ref{fig:specmrl}, the orthogonal basis update involves only one matrix multiplication and one QR factorization~\citep{qr}.
Counting a multiply-add as two FLOPs, the matrix multiplication requires $2D^3$ FLOPs, while QR factorization costs $\frac{8}{3}D^3$ FLOPs in total: $\frac{4}{3}D^3$ for factorization and another $\frac{4}{3}D^3$ for explicitly forming the full orthogonal matrix.
The additional forward cost is therefore given as:
\begin{equation}
\label{eq:extra-flops}
F_{\mathrm{extra}}
\approx
\underbrace{2RD^2}_{\text{Gram construction}}
+
\underbrace{\frac{14}{3}D^3}_{\text{basis update}}
+
\underbrace{2B_{\mathrm{proj}}D^2}_{\text{embedding projection}}
.
\end{equation}
For comparison, let $P$ denote the number of linear weights, and $T$ the number of tokens processed per device.
Following the approximation in~\citep{kaplan2020scaling}, the backbone forward cost is:
\begin{equation}
    \label{eq:backbone-flops}
    F_{\mathrm{backbone}}
    \approx
    2\,TP
    .
\end{equation}
Substituting the Qwen3-0.6B model configuration, the additional forward cost remains below 1\% of the backbone forward FLOPs, demonstrating the low overhead of our power-iteration design.
This computational efficiency is further validated by the empirical results presented in Section~\ref{sec:efficiency}.

\begin{table}[h]
    \centering
    \small
    \caption{MRL training configurations for different backbone models.}
    \label{tab:mrl_setup}
    \resizebox{0.86\textwidth}{!}{
        \begin{tabular}{l|ccccc}
            \toprule
             & \multicolumn{5}{c}{\textbf{MRL Setup}} \\
            \cmidrule(lr){1-1} \cmidrule(lr){2-6}
            \textrm{Checkpoint~[HF]} & $m_0~(w_0)$ & $m_1~(w_1)$ & $m_2~(w_2)$ & $m_3~(w_3)$ & $m_4~(w_4)$ \\
            \midrule
            \textrm{Qwen3-0.6B-Base}~\hflink{https://huggingface.co/Qwen/Qwen3-0.6B-Base} & $1024$\,\mrlweight{1.0} & $\padzero512$\,\mrlweight{0.3} & $256$\,\mrlweight{0.2} & $128$\,\mrlweight{0.1} & $\padzero64$\,\mrlweight{0.1} \\
            \textrm{Qwen2-0.5B}~\hflink{https://huggingface.co/Qwen/Qwen2-0.5B} & $\padzero896$\,\mrlweight{1.0} & $\padzero512$\,\mrlweight{0.3} & $256$\,\mrlweight{0.2} & $128$\,\mrlweight{0.1} & $\padzero64$\,\mrlweight{0.1} \\
            \textrm{Llama-3.2-1B}~\hflink{https://huggingface.co/meta-llama/Llama-3.2-1B} & $2048$\,\mrlweight{1.0} & $1024$\,\mrlweight{0.3} & $512$\,\mrlweight{0.2} & $256$\,\mrlweight{0.1} & $128$\,\mrlweight{0.1} \\
            \textrm{Qwen3-4B-Base}~\hflink{https://huggingface.co/Qwen/Qwen3-4B-Base} & $2560$\,\mrlweight{1.0} & $1024$\,\mrlweight{0.3} & $512$\,\mrlweight{0.2} & $256$\,\mrlweight{0.1} & $128$\,\mrlweight{0.1} \\
            \bottomrule
        \end{tabular}
    }
\end{table}

\begin{table}[htbp]
    \caption{MTEB evaluation metrics and the 15-dataset subset used for scaling experiments.}
    \label{tab:mteb_setup}
    \centering
    \small
    \setlength{\tabcolsep}{6pt}
    \renewcommand{\arraystretch}{1.10}

    \begin{tabular}{lll}
        \toprule
        \textbf{Category} & \textbf{Metric} & \textbf{Dataset} \\
        \midrule

        \multirow{3}{*}{STS. (3)}
        & \multirow{3}{*}{Spearman} & STS17~\citep{sts17} \\
        & & SICK-R~\citep{sickr} \\
        & & STSBenchmark~\citep{stsb} \\
        \midrule

        \multirow{3}{*}{Class. (3)}
        & \multirow{3}{*}{Accuracy} & Banking77Classification~\citep{banking77} \\
        & & EmotionClassification~\citep{emotion} \\
        & & MassiveIntentClassification~\citep{massive} \\
        \midrule

        \multirow{3}{*}{Cluster. (3)}
        & \multirow{3}{*}{V-measure} & \href{https://huggingface.co/datasets/mteb/biorxiv-clustering-s2s}{BiorxivClusteringS2S}~\citep{mteb} \\
        & & \href{https://huggingface.co/datasets/mteb/medrxiv-clustering-s2s}{MedrxivClusteringS2S}~\citep{mteb} \\
        & & TwentyNewsgroupsClustering~\citep{20newsgroups} \\
        \midrule

        PairClass. (1) & AP & SprintDuplicateQuestions~\citep{SprintDuplicateQuestions} \\
        \midrule

        \multirow{2}{*}{Rerank. (2)}
        & \multirow{2}{*}{MAP} & SciDocsRR~\citep{SciDocsRR} \\
        & & StackOverflowDupQuestions~\citep{StackOverflowDupQuestions} \\
        \midrule

        \multirow{3}{*}{Retr. (3)}
        & \multirow{3}{*}{nDCG@10} & SciFact~\citep{scifact} \\
        & & ArguAna~\citep{arguana} \\
        & & NFCorpus~\citep{nfcorpus} \\
        \bottomrule
    \end{tabular}
\end{table}

\begin{table*}[t]
    \caption{Task-level breakdown of MTEB results in Table~\ref{tab:main}. Shaded rows denote \name, bold values indicate better overall results, parenthetical values report gains over MRL under each setup.}
    \label{tab:breakdown}
    \centering
    \small
    \vskip 0.1in
    \renewcommand{\arraystretch}{1.18}
    \resizebox{0.96\textwidth}{!}{
        \begin{tabular}{c|ccccccc|c}
            \toprule
            \multicolumn{1}{c}{} & \textbf{\textrm{STS.}} & \textbf{\textrm{Class.}} & \textbf{\textrm{Cluster.}} & \textbf{\textrm{PairClass.}} & \textbf{\textrm{Rerank.}} & \textbf{\textrm{Retr.}} & \textbf{\textrm{Sum.}} & \textbf{\textrm{Avg}} \\
            \cmidrule(l{1pt}r{2pt}){1-1} \cmidrule(l{2pt}r{2pt}){2-8} \cmidrule(l{1pt}r{1pt}){9-9}
            \textbf{\textrm{\# Datasets }} & \small 10 & \small 12 & \small 11 & \small 3 & \small 4 & \small 15 & \small 1 & \small 56 \\
            \midrule

            & \multicolumn{8}{c}{\textbf{\textit{\textrm{Qwen3-0.6B}}}} \\
            \cmidrule(l{1pt}r{2pt}){1-1} \cmidrule(l{2pt}r{2pt}){2-8} \cmidrule(l{1pt}r{1pt}){9-9}

            \multirow{2}*[-2pt]{\shortstack{\textbf{1024} \\ \textbf{($\tau=1$)}}} & 80.77 & 74.25 & 48.72 & 72.74 & 48.93 & 49.02 & 30.38 & \textbf{60.97} \\
            \cmidrule(l{1pt}r{2pt}){2-9}
            & \cellcolor{gray!12}80.24 & \cellcolor{gray!12}74.45 & \cellcolor{gray!12}49.06 & \cellcolor{gray!12}72.44 & \cellcolor{gray!12}47.67 & \cellcolor{gray!12}49.30 & \cellcolor{gray!12}29.66 & \cellcolor{gray!12}60.94~($-$0.03) \\
            \cmidrule(l{1pt}r{2pt}){1-1} \cmidrule(l{2pt}r{2pt}){2-8} \cmidrule(l{1pt}r{1pt}){9-9}

            \multirow{2}*[-2pt]{\shortstack{\textbf{512} \\ \textbf{($\tau=2$)}}} & 80.87 & 73.99 & 48.43 & 72.09 & 48.25 & 48.10 & 29.92 & 60.54 \\
            \cmidrule(l{1pt}r{2pt}){2-9}
            & \cellcolor{gray!12}80.83 & \cellcolor{gray!12}75.11 & \cellcolor{gray!12}49.43 & \cellcolor{gray!12}74.49 & \cellcolor{gray!12}50.98 & \cellcolor{gray!12}49.24 & \cellcolor{gray!12}30.46 & \cellcolor{gray!12}\textbf{61.61~(+1.07)} \\
            \cmidrule(l{1pt}r{2pt}){1-1} \cmidrule(l{2pt}r{2pt}){2-8} \cmidrule(l{1pt}r{1pt}){9-9}

            \multirow{2}*[-2pt]{\shortstack{\textbf{256} \\ \textbf{($\tau=4$)}}} & 80.21 & 73.45 & 47.89 & 71.42 & 47.11 & 45.62 & 29.00 & 59.40 \\
            \cmidrule(l{1pt}r{2pt}){2-9}
            & \cellcolor{gray!12}81.04 & \cellcolor{gray!12}74.98 & \cellcolor{gray!12}49.44 & \cellcolor{gray!12}74.46 & \cellcolor{gray!12}50.75 & \cellcolor{gray!12}48.08 & \cellcolor{gray!12}30.30 & \cellcolor{gray!12}\textbf{61.28~(+1.88)} \\
            \cmidrule(l{1pt}r{2pt}){1-1} \cmidrule(l{2pt}r{2pt}){2-8} \cmidrule(l{1pt}r{1pt}){9-9}

            \multirow{2}*[-2pt]{\shortstack{\textbf{128} \\ \textbf{($\tau=8$)}}} & 79.49 & 72.99 & 47.43 & 68.29 & 45.47 & 42.18 & 29.24 & 57.88 \\
            \cmidrule(l{1pt}r{2pt}){2-9}
            & \cellcolor{gray!12}80.76 & \cellcolor{gray!12}74.56 & \cellcolor{gray!12}49.28 & \cellcolor{gray!12}73.90 & \cellcolor{gray!12}49.95 & \cellcolor{gray!12}45.93 & \cellcolor{gray!12}30.74 & \cellcolor{gray!12}\textbf{60.46~(+2.58)} \\
            \cmidrule(l{1pt}r{2pt}){1-1} \cmidrule(l{2pt}r{2pt}){2-8} \cmidrule(l{1pt}r{1pt}){9-9}

            \multirow{2}*[-2pt]{\shortstack{\textbf{64} \\ \textbf{($\tau=16$)}}} & 78.09 & 72.11 & 46.54 & 63.87 & 42.52 & 34.38 & 29.63 & 54.73 \\
            \cmidrule(l{1pt}r{2pt}){2-9}
            & \cellcolor{gray!12}80.18 & \cellcolor{gray!12}73.89 & \cellcolor{gray!12}48.89 & \cellcolor{gray!12}72.43 & \cellcolor{gray!12}48.07 & \cellcolor{gray!12}40.79 & \cellcolor{gray!12}29.86 & \cellcolor{gray!12}\textbf{58.53~(+3.80)} \\

            \midrule

            & \multicolumn{8}{c}{\textbf{\textit{\textrm{Qwen2-0.5B}}}} \\
            \cmidrule(l{1pt}r{2pt}){1-1} \cmidrule(l{2pt}r{2pt}){2-8} \cmidrule(l{1pt}r{1pt}){9-9}

            \multirow{2}*[-2pt]{\shortstack{\textbf{896} \\ \textbf{($\tau=1$)}}} & 78.57 & 71.82 & 46.99 & 71.52 & 46.68 & 47.04 & 30.59 & \textbf{58.96} \\
            \cmidrule(l{1pt}r{2pt}){2-9}
            & \cellcolor{gray!12}78.72 & \cellcolor{gray!12}71.80 & \cellcolor{gray!12}46.59 & \cellcolor{gray!12}71.33 & \cellcolor{gray!12}46.24 & \cellcolor{gray!12}46.75 & \cellcolor{gray!12}27.87 & \cellcolor{gray!12}58.74~($-$0.22) \\
            \cmidrule(l{1pt}r{2pt}){1-1} \cmidrule(l{2pt}r{2pt}){2-8} \cmidrule(l{1pt}r{1pt}){9-9}

            \multirow{2}*[-2pt]{\shortstack{\textbf{512} \\ \textbf{($\tau=2$)}}} & 78.49 & 72.10 & 46.80 & 70.91 & 46.13 & 46.14 & 31.13 & 58.67 \\
            \cmidrule(l{1pt}r{2pt}){2-9}
            & \cellcolor{gray!12}79.25 & \cellcolor{gray!12}73.07 & \cellcolor{gray!12}47.05 & \cellcolor{gray!12}73.22 & \cellcolor{gray!12}48.98 & \cellcolor{gray!12}46.51 & \cellcolor{gray!12}28.31 & \cellcolor{gray!12}\textbf{59.44~(+0.77)} \\
            \cmidrule(l{1pt}r{2pt}){1-1} \cmidrule(l{2pt}r{2pt}){2-8} \cmidrule(l{1pt}r{1pt}){9-9}

            \multirow{2}*[-2pt]{\shortstack{\textbf{256} \\ \textbf{($\tau=4$)}}} & 78.41 & 72.39 & 46.82 & 71.23 & 46.49 & 44.57 & 30.18 & 58.32 \\
            \cmidrule(l{1pt}r{2pt}){2-9}
            & \cellcolor{gray!12}79.40 & \cellcolor{gray!12}72.94 & \cellcolor{gray!12}47.11 & \cellcolor{gray!12}73.07 & \cellcolor{gray!12}48.85 & \cellcolor{gray!12}45.64 & \cellcolor{gray!12}28.37 & \cellcolor{gray!12}\textbf{59.20~(+0.88)} \\
            \cmidrule(l{1pt}r{2pt}){1-1} \cmidrule(l{2pt}r{2pt}){2-8} \cmidrule(l{1pt}r{1pt}){9-9}

            \multirow{2}*[-2pt]{\shortstack{\textbf{128} \\ \textbf{($\tau=8$)}}} & 77.71 & 72.26 & 46.68 & 69.69 & 45.03 & 41.94 & 30.67 & 57.26 \\
            \cmidrule(l{1pt}r{2pt}){2-9}
            & \cellcolor{gray!12}79.29 & \cellcolor{gray!12}72.79 & \cellcolor{gray!12}46.96 & \cellcolor{gray!12}72.25 & \cellcolor{gray!12}48.53 & \cellcolor{gray!12}43.34 & \cellcolor{gray!12}28.52 & \cellcolor{gray!12}\textbf{58.44~(+1.18)} \\
            \cmidrule(l{1pt}r{2pt}){1-1} \cmidrule(l{2pt}r{2pt}){2-8} \cmidrule(l{1pt}r{1pt}){9-9}

            \multirow{2}*[-2pt]{\shortstack{\textbf{64} \\ \textbf{($\tau=16$)}}} & 76.41 & 71.86 & 46.42 & 67.69 & 43.74 & 36.48 & 30.47 & 55.23 \\
            \cmidrule(l{1pt}r{2pt}){2-9}
            & \cellcolor{gray!12}78.33 & \cellcolor{gray!12}72.37 & \cellcolor{gray!12}46.72 & \cellcolor{gray!12}70.00 & \cellcolor{gray!12}46.55 & \cellcolor{gray!12}38.72 & \cellcolor{gray!12}27.59 & \cellcolor{gray!12}\textbf{56.61~(+1.38)} \\

            \midrule

            & \multicolumn{8}{c}{\textbf{\textit{\textrm{Llama3.2-1B}}}} \\
            \cmidrule(l{1pt}r{2pt}){1-1} \cmidrule(l{2pt}r{2pt}){2-8} \cmidrule(l{1pt}r{1pt}){9-9}

            \multirow{2}*[-2pt]{\shortstack{\textbf{2048} \\ \textbf{($\tau=1$)}}} & 80.91 & 75.77 & 50.76 & 75.43 & 47.71 & 49.60 & 29.64 & \textbf{61.92} \\
            \cmidrule(l{1pt}r{2pt}){2-9}
            & \cellcolor{gray!12}80.71 & \cellcolor{gray!12}76.05 & \cellcolor{gray!12}50.28 & \cellcolor{gray!12}76.02 & \cellcolor{gray!12}48.19 & \cellcolor{gray!12}49.39 & \cellcolor{gray!12}30.18 & \cellcolor{gray!12}61.87~($-$0.05) \\
            \cmidrule(l{1pt}r{2pt}){1-1} \cmidrule(l{2pt}r{2pt}){2-8} \cmidrule(l{1pt}r{1pt}){9-9}

            \multirow{2}*[-2pt]{\shortstack{\textbf{1024} \\ \textbf{($\tau=2$)}}} & 80.79 & 75.84 & 50.75 & 75.22 & 46.97 & 49.17 & 30.06 & 61.74 \\
            \cmidrule(l{1pt}r{2pt}){2-9}
            & \cellcolor{gray!12}81.09 & \cellcolor{gray!12}76.61 & \cellcolor{gray!12}50.60 & \cellcolor{gray!12}76.70 & \cellcolor{gray!12}51.27 & \cellcolor{gray!12}49.53 & \cellcolor{gray!12}30.35 & \cellcolor{gray!12}\textbf{62.42~(+0.68)} \\
            \cmidrule(l{1pt}r{2pt}){1-1} \cmidrule(l{2pt}r{2pt}){2-8} \cmidrule(l{1pt}r{1pt}){9-9}

            \multirow{2}*[-2pt]{\shortstack{\textbf{512} \\ \textbf{($\tau=4$)}}} & 80.55 & 75.73 & 50.64 & 74.77 & 46.63 & 48.30 & 29.40 & 61.36 \\
            \cmidrule(l{1pt}r{2pt}){2-9}
            & \cellcolor{gray!12}81.22 & \cellcolor{gray!12}76.59 & \cellcolor{gray!12}50.67 & \cellcolor{gray!12}76.62 & \cellcolor{gray!12}51.06 & \cellcolor{gray!12}48.80 & \cellcolor{gray!12}30.05 & \cellcolor{gray!12}\textbf{62.23~(+0.87)} \\
            \cmidrule(l{1pt}r{2pt}){1-1} \cmidrule(l{2pt}r{2pt}){2-8} \cmidrule(l{1pt}r{1pt}){9-9}

            \multirow{2}*[-2pt]{\shortstack{\textbf{256} \\ \textbf{($\tau=8$)}}} & 80.31 & 75.40 & 50.49 & 74.00 & 45.30 & 46.59 & 29.54 & 60.62 \\
            \cmidrule(l{1pt}r{2pt}){2-9}
            & \cellcolor{gray!12}81.28 & \cellcolor{gray!12}76.52 & \cellcolor{gray!12}50.53 & \cellcolor{gray!12}76.29 & \cellcolor{gray!12}50.57 & \cellcolor{gray!12}47.52 & \cellcolor{gray!12}30.19 & \cellcolor{gray!12}\textbf{61.80~(+1.18)} \\
            \cmidrule(l{1pt}r{2pt}){1-1} \cmidrule(l{2pt}r{2pt}){2-8} \cmidrule(l{1pt}r{1pt}){9-9}

            \multirow{2}*[-2pt]{\shortstack{\textbf{128} \\ \textbf{($\tau=16$)}}} & 79.88 & 74.79 & 50.34 & 73.20 & 44.88 & 43.21 & 30.29 & 59.42 \\
            \cmidrule(l{1pt}r{2pt}){2-9}
            & \cellcolor{gray!12}81.32 & \cellcolor{gray!12}76.28 & \cellcolor{gray!12}50.48 & \cellcolor{gray!12}75.40 & \cellcolor{gray!12}49.69 & \cellcolor{gray!12}44.74 & \cellcolor{gray!12}30.91 & \cellcolor{gray!12}\textbf{60.91~(+1.49)} \\

            \bottomrule
        \end{tabular}
    }
\end{table*}

\begin{table*}[t]
    \caption{Task-level breakdown of FeatMRL variants. Bold and underlined average scores indicate the best and second-best FeatMRL variants within each group.}
    \label{tab:featmrl-breakdown}
    \centering
    \small
    \vskip 0.1in
    \renewcommand{\arraystretch}{1.20}
    \resizebox{\textwidth}{!}{
        \begin{tabular}{l|ccccccc|c}
            \toprule
            & \textbf{\textrm{STS.}} & \textbf{\textrm{Class.}} & \textbf{\textrm{Cluster.}} & \textbf{\textrm{PairClass.}} & \textbf{\textrm{Rerank.}} & \textbf{\textrm{Retr.}} & \textbf{\textrm{Sum.}} & \textbf{\textrm{Avg}} \\
            \cmidrule(l{1pt}r{2pt}){1-1} \cmidrule(l{2pt}r{2pt}){2-8} \cmidrule(l{1pt}r{1pt}){9-9}
            \textbf{\textrm{\# Datasets}} & \small 10 & \small 12 & \small 11 & \small 3 & \small 4 & \small 15 & \small 1 & \small 56 \\
            \midrule

            $\mathbf{\tau=1}$ & 80.24 & 74.45 & 49.06 & 72.44 & 47.67 & 49.30 & 29.66 & 60.94 \\
            \cmidrule(l{1pt}r{2pt}){1-1} \cmidrule(l{2pt}r{2pt}){2-8} \cmidrule(l{1pt}r{1pt}){9-9}
            $\,\beta=0$ & 80.70 & 73.08 & 48.86 & 72.31 & 49.35 & 48.99 & 29.80 & \textbf{60.72}~($-$0.22) \\
            \midrule
            $\,\beta=0.1$ & 80.73 & 72.54 & 48.77 & 72.27 & 49.82 & 49.10 & 31.06 & 60.68~($-$0.26) \\
            $\,\beta=0.5$ & 80.65 & 72.30 & 48.79 & 72.87 & 49.54 & 49.29 & 31.65 & \underline{60.69}~($-$0.25) \\
            $\,\beta=0.9$ & 80.78 & 71.97 & 48.85 & 72.37 & 48.71 & 48.72 & 29.69 & 60.38~($-$0.56) \\

            \midrule

            $\mathbf{\tau=2}$ & 80.83 & 75.11 & 49.43 & 74.49 & 50.98 & 49.24 & 30.46 & 61.61 \\
            \cmidrule(l{1pt}r{2pt}){1-1} \cmidrule(l{2pt}r{2pt}){2-8} \cmidrule(l{1pt}r{1pt}){9-9}
            $\,\beta=0$ & 80.85 & 72.86 & 48.79 & 71.78 & 48.62 & 48.43 & 29.41 & \textbf{60.45}~($-$1.16) \\
            \midrule
            $\,\beta=0.1$ & 80.86 & 72.37 & 48.68 & 71.72 & 49.22 & 48.31 & 30.96 & 60.36~($-$1.25) \\
            $\,\beta=0.5$ & 80.79 & 71.98 & 48.56 & 72.45 & 48.96 & 48.85 & 31.85 & \underline{60.42}~($-$1.19) \\
            $\,\beta=0.9$ & 80.97 & 71.69 & 48.69 & 71.77 & 47.89 & 47.92 & 29.47 & 60.01~($-$1.60) \\

            \midrule

            $\mathbf{\tau=4}$ & 81.04 & 74.98 & 49.44 & 74.46 & 50.75 & 48.08 & 30.30 & 61.28 \\
            \cmidrule(l{1pt}r{2pt}){1-1} \cmidrule(l{2pt}r{2pt}){2-8} \cmidrule(l{1pt}r{1pt}){9-9}
            $\,\beta=0$ & 80.98 & 72.73 & 48.66 & 70.87 & 47.60 & 47.28 & 29.31 & \textbf{59.99}~($-$1.29) \\
            \midrule
            $\,\beta=0.1$ & 80.98 & 71.98 & 48.46 & 70.98 & 48.32 & 47.20 & 29.82 & 59.84~($-$1.44) \\
            $\,\beta=0.5$ & 80.79 & 71.69 & 48.51 & 71.68 & 48.02 & 47.66 & 31.25 & \underline{59.91}~($-$1.37) \\
            $\,\beta=0.9$ & 80.99 & 71.32 & 48.48 & 70.76 & 46.82 & 46.34 & 30.29 & 59.35~($-$1.93) \\

            \midrule

            $\mathbf{\tau=8}$ & 80.76 & 74.56 & 49.28 & 73.90 & 49.95 & 45.93 & 30.74 & 60.46 \\
            \cmidrule(l{1pt}r{2pt}){1-1} \cmidrule(l{2pt}r{2pt}){2-8} \cmidrule(l{1pt}r{1pt}){9-9}
            $\,\beta=0$ & 80.54 & 72.27 & 48.47 & 68.55 & 45.75 & 44.56 & 30.96 & \textbf{58.82}~($-$1.64) \\
            \midrule
            $\,\beta=0.1$ & 80.49 & 71.67 & 48.17 & 68.77 & 46.44 & 44.58 & 30.41 & 58.68~($-$1.78) \\
            $\,\beta=0.5$ & 80.39 & 71.21 & 48.22 & 69.37 & 46.24 & 45.12 & 31.54 & \underline{58.75}~($-$1.71) \\
            $\,\beta=0.9$ & 80.45 & 70.89 & 48.24 & 68.63 & 44.96 & 43.54 & 31.35 & 58.14~($-$2.32) \\

            \midrule

            $\mathbf{\tau=16}$ & 80.18 & 73.89 & 48.89 & 72.43 & 48.07 & 40.79 & 29.86 & 58.53 \\
            \cmidrule(l{1pt}r{2pt}){1-1} \cmidrule(l{2pt}r{2pt}){2-8} \cmidrule(l{1pt}r{1pt}){9-9}
            $\,\beta=0$ & 78.85 & 71.60 & 47.86 & 63.11 & 42.51 & 37.47 & 30.86 & \underline{55.83}~($-$2.70) \\
            \midrule
            $\,\beta=0.1$ & 78.73 & 70.95 & 47.58 & 63.54 & 43.23 & 37.26 & 29.82 & 55.62~($-$2.91) \\
            $\,\beta=0.5$ & 78.71 & 70.78 & 47.76 & 63.82 & 43.15 & 38.35 & 29.59 & \textbf{55.90}~($-$2.63) \\
            $\,\beta=0.9$ & 78.82 & 70.28 & 47.79 & 63.63 & 41.91 & 36.18 & 30.12 & 55.15~($-$3.38) \\

            \bottomrule
        \end{tabular}
    }
\end{table*}

\begin{table*}[t]
    \caption{Task-level breakdown of MTEB in an over-training regime on Qwen3-0.6B, with training steps extended from 3K to 15K. The improvement of \name\ remains consistent with Table~\ref{tab:main}, suggesting that the gains of \name\ persist beyond faster convergence.}
    \label{tab:allstep-breakdown}
    \centering
    \small
    \renewcommand{\arraystretch}{1.18}
    \resizebox{0.90\textwidth}{!}{
        \begin{tabular}{c|ccccccc|c}
            \toprule
            \multicolumn{1}{c}{}
            & \textbf{\textrm{STS.}}
            & \textbf{\textrm{Class.}}
            & \textbf{\textrm{Cluster.}}
            & \textbf{\textrm{PairClass.}}
            & \textbf{\textrm{Rerank.}}
            & \textbf{\textrm{Retr.}}
            & \textbf{\textrm{Sum.}}
            & \textbf{\textrm{Avg}} \\
            \cmidrule(l{1pt}r{2pt}){1-1}
            \cmidrule(l{2pt}r{2pt}){2-8}
            \cmidrule(l{1pt}r{1pt}){9-9}

            \textbf{\textrm{\# Datasets}}
            & \small 10
            & \small 12
            & \small 11
            & \small 3
            & \small 4
            & \small 15
            & \small 1
            & \small 56 \\
            \midrule

            \multirow{2}*[-2pt]{
                \shortstack{\textbf{1024}\\\textbf{($\tau=1$)}}
            }
            & 80.52
            & 81.40
            & 52.79
            & 70.19
            & 48.99
            & 50.69
            & 28.91
            & \textbf{63.54} \\
            \cmidrule(l{1pt}r{2pt}){2-9}
            & \cellcolor{gray!12}80.74
            & \cellcolor{gray!12}81.15
            & \cellcolor{gray!12}52.36
            & \cellcolor{gray!12}68.12
            & \cellcolor{gray!12}47.86
            & \cellcolor{gray!12}50.59
            & \cellcolor{gray!12}30.38
            & \cellcolor{gray!12}63.25~(-0.29) \\

            \cmidrule(l{1pt}r{2pt}){1-1}
            \cmidrule(l{2pt}r{2pt}){2-8}
            \cmidrule(l{1pt}r{1pt}){9-9}

            \multirow{2}*[-2pt]{
                \shortstack{\textbf{512}\\\textbf{($\tau=2$)}}
            }
            & 80.53
            & 81.34
            & 52.74
            & 69.44
            & 48.43
            & 49.80
            & 28.28
            & 63.19 \\
            \cmidrule(l{1pt}r{2pt}){2-9}
            & \cellcolor{gray!12}81.35
            & \cellcolor{gray!12}81.50
            & \cellcolor{gray!12}52.64
            & \cellcolor{gray!12}71.39
            & \cellcolor{gray!12}50.95
            & \cellcolor{gray!12}50.59
            & \cellcolor{gray!12}29.75
            & \cellcolor{gray!12}\textbf{63.88~(+0.69)} \\

            \cmidrule(l{1pt}r{2pt}){1-1}
            \cmidrule(l{2pt}r{2pt}){2-8}
            \cmidrule(l{1pt}r{1pt}){9-9}

            \multirow{2}*[-2pt]{
                \shortstack{\textbf{256}\\\textbf{($\tau=4$)}}
            }
            & 80.20
            & 81.12
            & 52.49
            & 68.01
            & 46.83
            & 47.52
            & 27.82
            & 62.23 \\
            \cmidrule(l{1pt}r{2pt}){2-9}
            & \cellcolor{gray!12}81.55
            & \cellcolor{gray!12}81.41
            & \cellcolor{gray!12}52.60
            & \cellcolor{gray!12}71.44
            & \cellcolor{gray!12}50.85
            & \cellcolor{gray!12}49.69
            & \cellcolor{gray!12}29.50
            & \cellcolor{gray!12}\textbf{63.64~(+1.41)} \\

            \cmidrule(l{1pt}r{2pt}){1-1}
            \cmidrule(l{2pt}r{2pt}){2-8}
            \cmidrule(l{1pt}r{1pt}){9-9}

            \multirow{2}*[-2pt]{
                \shortstack{\textbf{128}\\\textbf{($\tau=8$)}}
            }
            & 79.50
            & 80.91
            & 52.28
            & 64.48
            & 45.29
            & 44.91
            & 28.53
            & 61.03 \\
            \cmidrule(l{1pt}r{2pt}){2-9}
            & \cellcolor{gray!12}81.23
            & \cellcolor{gray!12}81.09
            & \cellcolor{gray!12}52.52
            & \cellcolor{gray!12}70.75
            & \cellcolor{gray!12}50.42
            & \cellcolor{gray!12}47.33
            & \cellcolor{gray!12}28.13
            & \cellcolor{gray!12}\textbf{62.77~(+1.74)} \\

            \cmidrule(l{1pt}r{2pt}){1-1}
            \cmidrule(l{2pt}r{2pt}){2-8}
            \cmidrule(l{1pt}r{1pt}){9-9}

            \multirow{2}*[-2pt]{
                \shortstack{\textbf{64}\\\textbf{($\tau=16$)}}
            }
            & 78.21
            & 80.29
            & 51.88
            & 61.40
            & 43.08
            & 39.03
            & 28.71
            & 58.69 \\
            \cmidrule(l{1pt}r{2pt}){2-9}
            & \cellcolor{gray!12}80.45
            & \cellcolor{gray!12}80.23
            & \cellcolor{gray!12}52.28
            & \cellcolor{gray!12}69.22
            & \cellcolor{gray!12}49.26
            & \cellcolor{gray!12}42.12
            & \cellcolor{gray!12}28.85
            & \cellcolor{gray!12}\textbf{60.85~(+2.16)} \\

            \bottomrule
        \end{tabular}
    }
\end{table*}

\begin{table*}[t]
    \caption{
    Task-level breakdown of MTEB results for \name\ and existing MRL variants. Column $\mathrm{Avg}_{\mathrm{w/o\ Cls}}$ denotes the average MTEB results over non-classification MTEB datasets.
    }
    \label{tab:mrl-improvements-breakdown}
    \centering
    \small
    \vskip 0.1in
    \renewcommand{\arraystretch}{1.15}
    \resizebox{\textwidth}{!}{
        \begin{tabular}{c|c|ccccccc|c}
            \toprule
            & \textbf{Method} & \textbf{\textrm{STS.}} & \textbf{\textrm{Class.}} & \textbf{\textrm{Cluster.}} & \textbf{\textrm{PairClass.}} & \textbf{\textrm{Rerank.}} & \textbf{\textrm{Retr.}} & \textbf{\textrm{Sum.}} & $\mathrm{Avg}_{\mathrm{w/o\,Cls}}$ \\
            \cmidrule(l{1pt}r{2pt}){1-2} \cmidrule(l{2pt}r{2pt}){3-9} \cmidrule(l{1pt}r{1pt}){10-10}

            \multirow{4}{*}{\shortstack{\textbf{1024}\\$\mathbf{(\tau=1)}$}} & MRL & 80.77 & 74.25 & 48.72 & 72.74 & 48.93 & 49.02 & 30.38 & 57.35 \\
            \cmidrule(l{2pt}r{2pt}){2-9} \cmidrule(l{1pt}r{1pt}){10-10}
            & ESE & 81.12 & 59.88 & 48.61 & 70.96 & 49.12 & 49.49 & 29.93 & \textbf{57.44}~($+$0.10) \\
            & MIPIC & 79.88 & 73.05 & 48.75 & 69.69 & 48.51 & 49.21 & 30.77 & 56.98~($-$0.37) \\
            & \cellcolor{gray!12}\name & \cellcolor{gray!12}80.24 & \cellcolor{gray!12}74.45 & \cellcolor{gray!12}49.06 & \cellcolor{gray!12}72.44 & \cellcolor{gray!12}47.67 & \cellcolor{gray!12}49.30 & \cellcolor{gray!12}29.66 & \cellcolor{gray!12}57.26~($-$0.09) \\

            \midrule

            \multirow{4}{*}{\shortstack{\textbf{512}\\$\mathbf{(\tau=2)}$}} & MRL & 80.87 & 73.99 & 48.43 & 72.09 & 48.25 & 48.10 & 29.92 & 56.87 \\
            \cmidrule(l{2pt}r{2pt}){2-9} \cmidrule(l{1pt}r{1pt}){10-10}
            & ESE & 80.92 & 59.12 & 48.47 & 69.47 & 48.49 & 48.23 & 28.85 & 56.75~($-$0.12) \\
            & MIPIC & 80.02 & 72.82 & 48.56 & 68.49 & 47.94 & 48.03 & 30.12 & 56.41~($-$0.45) \\
            & \cellcolor{gray!12}\name & \cellcolor{gray!12}80.83 & \cellcolor{gray!12}75.11 & \cellcolor{gray!12}49.43 & \cellcolor{gray!12}74.49 & \cellcolor{gray!12}50.98 & \cellcolor{gray!12}49.24 & \cellcolor{gray!12}30.46 & \cellcolor{gray!12}\textbf{57.92}~($+$1.05) \\

            \midrule

            \multirow{4}{*}{\shortstack{\textbf{256}\\$\mathbf{(\tau=4)}$}} & MRL & 80.21 & 73.45 & 47.89 & 71.42 & 47.11 & 45.62 & 29.00 & 55.57 \\
            \cmidrule(l{2pt}r{2pt}){2-9} \cmidrule(l{1pt}r{1pt}){10-10}
            & ESE & 80.40 & 58.10 & 48.17 & 71.04 & 48.07 & 46.06 & 30.02 & 55.91~($+$0.35) \\
            & MIPIC & 79.87 & 72.47 & 48.23 & 67.35 & 46.39 & 46.03 & 31.57 & 55.43~($-$0.14) \\
            & \cellcolor{gray!12}\name & \cellcolor{gray!12}81.04 & \cellcolor{gray!12}74.98 & \cellcolor{gray!12}49.44 & \cellcolor{gray!12}74.46 & \cellcolor{gray!12}50.75 & \cellcolor{gray!12}48.08 & \cellcolor{gray!12}30.30 & \cellcolor{gray!12}\textbf{57.55}~($+$1.98) \\

            \midrule

            \multirow{4}{*}{\shortstack{\textbf{128}\\$\mathbf{(\tau=8)}$}} & MRL & 79.49 & 72.99 & 47.43 & 68.29 & 45.47 & 42.18 & 29.24 & 53.76 \\
            \cmidrule(l{2pt}r{2pt}){2-9} \cmidrule(l{1pt}r{1pt}){10-10}
            & ESE & 79.79 & 57.19 & 47.81 & 68.43 & 47.33 & 43.76 & 29.12 & 54.63~($+$0.88) \\
            & MIPIC & 79.36 & 72.07 & 47.96 & 63.52 & 44.77 & 43.13 & 30.51 & 53.82~($+$0.07) \\
            & \cellcolor{gray!12}\name & \cellcolor{gray!12}80.76 & \cellcolor{gray!12}74.56 & \cellcolor{gray!12}49.28 & \cellcolor{gray!12}73.90 & \cellcolor{gray!12}49.95 & \cellcolor{gray!12}45.93 & \cellcolor{gray!12}30.74 & \cellcolor{gray!12}\textbf{56.61}~($+$2.85) \\

            \midrule

            \multirow{4}{*}{\shortstack{\textbf{64}\\$\mathbf{(\tau=16)}$}} & MRL & 78.09 & 72.11 & 46.54 & 63.87 & 42.52 & 34.38 & 29.63 & 50.00 \\
            \cmidrule(l{2pt}r{2pt}){2-9} \cmidrule(l{1pt}r{1pt}){10-10}
            & ESE & 77.59 & 54.63 & 46.45 & 61.87 & 43.13 & 32.43 & 27.75 & 49.07~($-$0.92) \\
            & MIPIC & 78.08 & 71.20 & 47.20 & 59.58 & 42.22 & 36.88 & 29.96 & 50.70~($+$0.70) \\
            & \cellcolor{gray!12}\name & \cellcolor{gray!12}80.18 & \cellcolor{gray!12}73.89 & \cellcolor{gray!12}48.89 & \cellcolor{gray!12}72.43 & \cellcolor{gray!12}48.07 & \cellcolor{gray!12}40.79 & \cellcolor{gray!12}29.86 & \cellcolor{gray!12}\textbf{54.34}~($+$4.34) \\

            \bottomrule
        \end{tabular}
    }
\end{table*}

\end{document}